\documentclass[electronic,journal]{abbrv_journal/vgtc}             % electronic version

\newcommand{\manuscriptnotetxt}{}

\graphicspath{{figures/}{pictures/}{images/}{./}} % where to search for the images
\PassOptionsToPackage{dvipsnames}{xcolor}
\usepackage{microtype}                 % use micro-typography (slightly more compact, better to read)
\PassOptionsToPackage{warn}{textcomp}  % to address font issues with \textrightarrow
\usepackage{textcomp}                  % use better special symbols
\usepackage{mathptmx}                  % use matching math font
\usepackage{times}                     % we use Times as the main font
\usepackage{cite}                      % needed to automatically sort the references
\usepackage{tabu}                      % only used for the table example
\usepackage{booktabs}                  % only used for the table example
\usepackage{lipsum}                    % used to generate placeholder text
\usepackage{mwe}                       % used to generate placeholder figures
\usepackage{bm}
\usepackage{algorithm}
\usepackage[algo2e,linesnumbered,ruled,vlined]{algorithm2e}
\usepackage{makecell}
\usepackage{amsmath,amssymb}
\usepackage{multirow}
\usepackage{float}
\usepackage{arydshln}
\usepackage{bbding}
\usepackage{pgfplots}
\usepackage{tikz}
\usetikzlibrary{shapes.geometric, arrows, positioning, fit, calc}
\usepackage{authblk}
\pgfplotsset{compat=1.18}

\tikzset{
    process/.style={
        rectangle, 
        draw=black, 
        fill=blue!20, 
        text width=5em, 
        text centered, 
        rounded corners, 
        minimum height=4em
    },
    arrow/.style={thick,->,>=stealth}
}
\vgtcpapertype{algorithm/technique}

\makeatletter
\def\thanks#1{\protected@xdef\@thanks{\@thanks
        \protect\footnotetext{#1}}}
\makeatother

\title{XR-VIO: High-precision Visual Inertial Odometry with Fast Initialization for XR Applications}

\author{Shangjin Zhai$^*$, Nan Wang$^*$, Xiaomeng Wang$^*$, Danpeng Chen, Weijian Xie, Hujun Bao, Guofeng Zhang$^\dagger$
\thanks{S. Zhai, N. Wang and X. Wang are with SenseTime Research. E-mails: zhaishangjin@sensetime.com, wangnan@sensetime.com, wangxiaomeng@sensetime.com.}
\thanks{W. Xie and D. Chen are with the State Key Lab of CAD\&CG, Zhejiang University and SenseTime Research. D. Chen is also affiliated with Tetras.AI. E-mails: xieweijian@sensetime.com, chendanpeng@tetras.ai.}
\thanks{H. Bao and G. Zhang are with the State Key Lab of CAD\&CG, Zhejiang University. E-mails: \{baohujun, zhangguofeng\}@zju.edu.cn.}
\thanks{$^*$ Equal Contribution}
\thanks{$^\dagger$ Corresponding\ Author}
}

\abstract {
This paper presents a novel approach to Visual Inertial Odometry (VIO), focusing on the initialization and feature matching modules. Existing methods for initialization often suffer from either poor stability in visual Structure from Motion (SfM) or fragility in solving a huge number of parameters simultaneously. To address these challenges, we propose a new pipeline for visual inertial initialization that robustly handles various complex scenarios. By tightly coupling gyroscope measurements, we enhance the robustness and accuracy of visual SfM. Our method demonstrates stable performance even with only four image frames, yielding competitive results.
In terms of feature matching, we introduce a hybrid method that combines optical flow and descriptor-based matching. By leveraging the robustness of continuous optical flow tracking and the accuracy of descriptor matching, our approach achieves efficient, accurate, and robust tracking results. 
Through evaluation on multiple benchmarks, our method demonstrates state-of-the-art performance in terms of accuracy and success rate. Additionally, a video demonstration on mobile devices showcases the practical applicability of our approach in the field of Augmented Reality/Virtual Reality (AR/VR).
} % end of abstract


\teaser{
  \centering
  \includegraphics[width=\linewidth ]{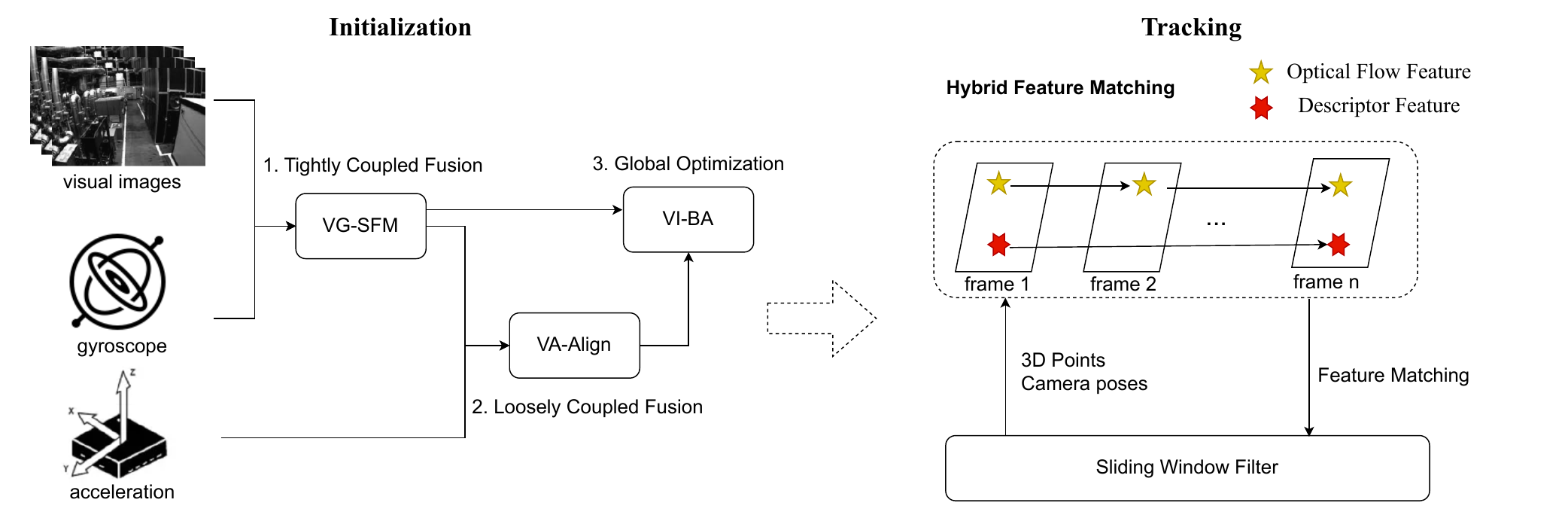}
  \caption{This work primarily consists of two key modules: initialization and feature matching. The left side illustrates our initialization scheme based on visual-inertial tightly coupled VG-SfM, resulting in substantial improvements in accuracy and initialization success rate. On the right side is our proposed hybrid matching strategy, which effectively integrates optical flow-based and descriptor-based matching methods, thereby significantly enhancing the accuracy of VIO.
  }
  \label{fig:teaser}
}

\begin{document}
\maketitle

\firstsection{Introduction}

%% \section{Introduction} %for journal use above \firstsection{..} instead
Visual Inertial Odometry (VIO) is a pivotal technology that integrates image and Inertial Measurement Unit (IMU) measurements to estimate the 6 degrees of freedom (6DoF) motion of a camera. Widely adopted in Augmented Reality/Virtual Reality (AR/VR) systems and autonomous navigation, VIO stands out for its utilization of low-cost and compact camera and IMU sensors. The performance of VIO systems heavily relies on visual inertial initialization and feature matching. Moreover, the robustness and low latency of initialization are critical for Extended Reality (XR) applications, while developers expect accurate camera tracking within milliseconds of launching VIO, regardless of the use case.

The initialization of VIO involves estimating initial variables such as gravity, velocity, and biases of gyroscope and accelerometer for sensors with calibrated intrinsic and extrinsic parameters. Accurate initialization of VIO algorithms is crucial for providing consistent and accurate motion tracking. 
    As shown in \cref{tab:algorithms}, early methods \cite{Dong-Si_initialization, martinelli2014closed} attempted to directly solve all initial variables by constructing a set of equations incorporating visual and IMU observations. Although these algorithms have low computational complexity, they lack robustness and are susceptible to outliers. 
    In recent years, some methods (e.g., \cite{qin-tro-2018_VINS-Mono,campos2020inertial,zuniga2021analytical,campos2019fast, Qin_Shen_2017}) have adopted a loosely coupled approach. They primarily rely on visual Structure from Motion (SfM) to reconstruct the initial structure and align it with IMU pre-integration\cite{forster2017manifold-preintergration}. Consequently, the quality of initialization highly depends on visual SfM. However, in cases of low parallax or small fragments, it becomes challenging to solve the problem robustly. 
    To address the robustness issue in VIO initialization, \cite{Rotation-Translation-Decoupled} proposed a rotation and translation decoupled method. This method first estimates rotation-related parameters and then employs a linear global translation constraint to solve other states without reconstructing 3D points, thereby increasing the success rate of initialization with considerable accuracy. However, this pose-only method is unable to fully exploit nonlinearities in the initialization problems, which limits the accuracy. 
    Furthermore, currently available consumer-grade IMU sensors can provide relatively accurate rotational measurements even without calibrated biases. These measurements can directly offer good initial rotation values for VIO system initialization, significantly simplifying the problem. However, previous algorithms have not fully utilized such characteristics.

\begin{table*}[htbp]
  \caption{Comparison of our method with previous VIO initialization approaches. "Gyr" denotes gyroscope and "Acc" denotes accelerometer. 
  "Decoupled R\&T 2D" refers to rotation and translation decoupling with  2d visual information, while "Decoupled R\&T 3D" refers to the same, but with 3D information.
Our method comprehensively leverages visual, IMU and 3D information. In contrast, existing VI tightly-coupled methods suffer from instability due to crude fusion of visual and IMU data; VI loosely-coupled methods do not integrate IMU data into the visual SfM process; Decoupled R\&T  2D methods neglect the use of 3D information to mitigate system errors. These shortcomings in existing methods significantly affect the accuracy, robustness and success rate of VIO initialization.}
  \centering
  
    % \begin{tabular}{c|c|c|c|c|c}
    % \toprule
    % \textbf{Algorithms}& \textbf{VI Coupling Mode} & \textbf{Visual SFM} & \textbf{Two View Reconstruction}& \textbf{Estimate IMU Bias}
    % & \textbf{VI-BA \& 3D Map} \\
    % \midrule
    % Closed-form \cite{martinelli2014closed} & Gyro \& Acc Tight & No & No & No
    % & \textbf{Yes} \\
    % VINS-Mono \cite{qin-tro-2018_VINS-Mono} & Gyro \& Acc Loose & Yes & 5-point Ransac & \textbf{Gyro} & \textbf{Yes} \\
    % Inertial-only \cite{campos2020inertial} & Gyro \& Acc Loose & Yes & 4-point \& 8-point Ransac & Gyro \& Acc  & \textbf{Yes} \\
    % DRT-t \cite{Rotation-Translation-Decoupled}  & Gyro \& Acc Tight & No & No &\textbf{Gyro}  & No \\
    % DRT-l \cite{Rotation-Translation-Decoupled}  & \textbf{Gyro Tight, Acc Loose} & No & No & \textbf{Gyro} & No \\
    % Ours  & \textbf{Gyro Tight, Acc Loose} & \textbf{VG-SFM} & \textbf{2-point Ransac w/ Gyro} & \textbf{Gyro}  & \textbf{Yes} \\
    % \end{tabular}

    \begin{tabular}{cccccccc}
    \toprule
    \textbf{VI Coupling Type} & \textbf{Algorithms} & \textbf{Visual Init}& \textbf{Visual SfM} & \textbf{VI-BA}  & \textbf{Bias Init}& \textbf{Robustness} & \textbf{Accuracy} \\
    \midrule
    Tightly coupled& Closed-form \cite{martinelli2014closed,geneva2020openvins} & - & - & \checkmark  &-& $\star\star$ & $\star\star$ \\
    \midrule
    \multirow{2}{*}{Loosely coupled} & Inertial-Only \cite{campos2020inertial} & 4-point/8-point& \checkmark & \checkmark  &Gyr \& Acc& $\star\star\star$  & $\star\star$ \\
    & VINS-Mono \cite{qin-tro-2018_VINS-Mono}& 5-point& \checkmark & \checkmark  &Gyr& $\star\star\star$ & $\star\star\star$ \\
    \midrule
    \multirow{2}{*}{Decoupled R\&T 2D}& DRT-t \cite{Rotation-Translation-Decoupled} & - & - & -  &Gyr& $\star\star\star\star$ & $\star\star\star$ \\
     & DRT-l \cite{Rotation-Translation-Decoupled}& - & - & -  &Gyr& $\star\star\star\star$ & $\star\star\star\star$ \\
    \midrule
    \textbf{Decoupled R\&T 3D}& XR-VIO& \textbf{2-point w/ Gyr}& \textbf{VG-SfM} & \textbf{\checkmark}  &Gyr& $\star\star\star\star\star$ & $\star\star\star\star\star$ \\
    \bottomrule
    \end{tabular}

  \label{tab:algorithms}
\end{table*}%
Conventional methods of VIO or visual SLAM typically utilize popular optical flow-based methods \cite{Lucas_Kanade_1981_KLT} or descriptors-based methods, such as BRIEF\cite{calonder2010brief}, ORB\cite{rublee2011orb}, and BRISK\cite{BRISK},  to match keypoint features and estimate the state. However, each of these techniques has its respective drawbacks: optical flow-based methods tend to drift after long feature tracking, while descriptor-based methods often encounter tracking failures, both resulting in inaccurate state estimation. To address these challenges, some studies have explored the utilization of geometric constraints, such as planarity or depth information. Additionally, other approaches have leveraged deep learning techniques, such as LoFTR\cite{sun2021loftr} and SuperGlue\cite{sarlin2020superglue}, to learn feature representations and matching strategies that are robust to textureless regions or repetitive patterns. Nevertheless, for XR and mobile applications, he complexity of visual features significantly increase the computational load. 

% \begin{table*}[h]
%   \centering
%     \begin{tabular}{c|c|c|c|c|c}
%     \toprule
%     \textbf{Algorithms}& \textbf{VI Coupling Mode} & \textbf{Visual SFM} & \textbf{Two View Reconstruction}& \textbf{Estimate IMU Bias}
%     & \textbf{VI-BA \& 3D Map} \\
%     \midrule
%     Closed-form \cite{martinelli2014closed} & Gyro \& Acc Tight & No & No & No
%     & \textbf{Yes} \\
%     VINS-Mono \cite{qin-tro-2018_VINS-Mono} & Gyro \& Acc Loose & Yes & Visual & \textbf{Gyro} & \textbf{Yes} \\
%     Inertial-only \cite{campos2020inertial} & Gyro \& Acc Loose & Yes & Visual & Gyro \& Acc  & \textbf{Yes} \\
%     DRT-t \cite{Rotation-Translation-Decoupled}  & Gyro \& Acc Tight & No & No &\textbf{Gyro}  & No \\
%     DRT-l \cite{Rotation-Translation-Decoupled}  & \textbf{Gyro Tight, Acc Loose} & No & No & \textbf{Gyro} & No \\
%     Ours  & \textbf{Gyro Tight, Acc Loose} & \textbf{VG-SFM} & \textbf{Visual \& Gyro} & \textbf{Gyro}  & \textbf{Yes} \\
%     \end{tabular}%
%   \label{tab:algorithms}%
% \end{table*}%
In this article, we aim to explore the limits of visual inertial initialization. Through a novel visual inertial fusion process, we significantly improve the success rate and accuracy of initialization. Even in short fragments with small parallax, our proposed initialization method can be completed quickly and stably. 
Additionally, we explore various feature matching solutions to achieve superior matching results by combining existing algorithms suitable for mobile platforms. 
By elucidating these advancements, this paper seeks to contribute to the development of more dependable and efficient VIO systems for XR, enabling immersive and realistic user experiences.

Our main contributions can be summarized as follows:
\begin{itemize}
\item \textbf{Fast VI Initialization}: We propose a rapid and precise VI initialization method, which robustly estimates the initial state of VIO using only four image frames. This robust estimation enables users to freely use mobile devices for AR experiences. In the EuRoC \cite{Burri25012016-EuRoC} benchmark evaluation, our algorithm achieves the state-of-the-art results compared to other similar algorithms.

\item \textbf{Hybrid Feature Matching}: We introduce a hybrid feature matching method, which tightly integrates two traditional feature matching schemes: optical flow and descriptor matching. By leveraging the advantages of both approaches, features can be tracked more stably and accurately. This results in longer track length and reduced feature drift, leading to more accurate state estimation of VIO.

\item \textbf{High-Precision VIO}: We develop a complete VIO system. With the improvements in initialization and feature matching, our entire system demonstrates outstanding accuracy and robustness. This system achieves state-of-the-art performance on open-source datasets compared to other feature matching-based VIO methods.
\end{itemize}
\section{Related Work}

Currently, there are numerous remarkable works on VIO in both academia and industry, which fall into two main categories: filtering-based and optimization-based approaches. Additionally, there are several typical implementations in the community for the two key modules of initialization and feature matching.

%% copy from RD-VIO as baseline %%
 \textbf{Filtering-based VIO} Early VIO systems, such as MSCKF \cite{mourikis-icra-2007} and ROVIO \cite{bloesch-ijrr-2017}, were built on Kalman filtering principles. These systems, based on sliding window approaches, incorporate frame poses at different time steps into their state vector.  During the update phase, visual constraints are utilized to update the state vector, while landmarks are marginalized. This approach limited the computational complexity required for these systems. ROVIO, in particular, integrated data association with the filter estimation process by utilizing photometric errors for visual observations. 
 
R-VIO \cite{huai2022robocentric} introduced a novel robocentric visual-inertial odometry approach. The key idea is to redesign the VINS framework based on a moving local coordinate system, rather than the fixed global reference frame typically used in conventional world-centric VINS. This redesign aims to achieve higher precision relative motion estimation for updating the global pose. 

OpenVINS \cite{geneva2020openvins} emerged as a recently developed open-source platform that employed the MSCKF filter. Its modular design allowed for flexibility of use and easy expansion. Evaluations on open-source datasets demonstrated its superior precision and robustness.

To boost VIO performance, recent systems, as discussed in \cite{bao2022robust}, have leveraged pre-existing high-precision maps, resulting in significant enhancements in accuracy. Others, such as RNIN-VIO \cite{chen2021rnin}, have capitalized on neural networks in IMU navigation to enhance robustness. These advancements showcase the ongoing efforts in pushing the boundaries of VIO.

\textbf{Optimization-based VIO }OKVIS \cite{leutenegger-ijrr-2015-OKVIS} is a system that operates using a sliding window methodology. During the optimization process, it integrates new keyframes and marginalizes old keyframes. By linearizing old observations as priors and incorporating them into the optimization, the system achieves improved performance. VINS-Mono \cite{qin-tro-2018_VINS-Mono} and VINS-Fusion \cite{qin2019a_VINS_Fusion_Local,qin2019b_VINS_Fusion_Global} are recent influential VI-SLAM systems that also use keyframe-based sliding window methods in their frontends. The backend loop-closure feature helps mitigate accumulated errors, resulting in higher precision. In contrast, VI-ORB-SLAM \cite{murartal-ral-2017-VI-ORB} is a loosely-coupled VI-SLAM system that initially processes visual observations using traditional VSLAM techniques before aligning them with inertial measurements to obtain metric results. ORB-SLAM3 \cite{campos2021orb-slam3} is a tightly-coupled system known for its remarkable accuracy. However, it solves the full SLAM problem, necessitating optimization of early poses based on later observations, making it unsuitable for real-time applications. VI-DSO \cite{von2018direct-VI-DSO} and DM-VIO \cite{stumberg22DM-VIO} are VIO extensions of the original DSO \cite{engel2018direct_DSO}, designed to enhance robustness and provide accurate scaling capabilities.

\textbf{Visual Inertial Initialization} The accurate and robust initialization of VIO is crucial for its normal operation. This process involves using both visual and inertial measurements to calculate the initial states, including scale, speed, gravity direction, etc. In the work of \cite{Dong-Si_initialization}, a tightly coupled closed-form algorithm was proposed, which uses visual and IMU measurements to simultaneously estimate initial states and the depth of feature points. OpenVINS \cite{geneva2020openvins}, a recently released open-source VIO platform, adopts this initialization process and adds Visual Inertial Bundle Adjustment \cite{triggs1999bundle} (VI-BA) after state initialization, as discussed in \cite{genevaopenvins}. On the other hand, VINS-Mono\cite{qin-tro-2018_VINS-Mono} is a typical loosely coupled method, which firstly utilizes visual SfM to solve camera poses using only visual measurements and then aligns them with IMU pre-intergration \cite{forster2017manifold-preintergration} to estimate the initial state. 
The latest approach \cite{Rotation-Translation-Decoupled} employs a decoupled formulation for rotation and translation, along with a pose-only solver, to estimate the initial state. Experiments in \cite{Rotation-Translation-Decoupled} have shown that it is optimal to firstly tightly couple the gyroscope in visual-inertial optimization, and secondly loosely couple the accelerometer in linear VI-alignment (namely DRT-l). Additionally, capitalizing on advancements in monocular depth estimation methods \cite{ranftl2020towards}, \cite{Mono-Depth-VI-Init-2022,merrill2023fast} employ the learned monocular depth as input for multi-view consistent visual-inertial initialization. We provide a detailed comparison of initialization algorithms in \cref{tab:algorithms}. 

\textbf{Feature Matching}
    GFTT\cite{gftt1994shi} and KLT\cite{Lucas_Kanade_1981_KLT} are popular optical flow-based feature matching methods for VIO, both of which are implemented in OpenCV \cite{opencv_library}. VINS-Mono\cite{qin-tro-2018_VINS-Mono}, OpenVINS\cite{geneva2020openvins} and many other VIO systems achieve high accuracy using these implementations.  Optical flow is suitable for consecutive feature matching and often results in long track length. However, it may suffer from frame-by-frame drift, posing a core bottleneck for VIO accuracy.  On the other hand, descriptor-based matching, as utilized by systems like ORB-SLAM\cite{leutenegger-ijrr-2015-OKVIS,campos2021orb-slam3,murartal-ral-2017-VI-ORB}, offers better accuracy and lower drift. BRISK\cite{BRISK} and ORB\cite{rublee2011orb} are two widely used methods for feature extraction and description, both known for their good repeatability and suitability for feature matching. Despite these advantages, descriptor matching tends to have a shorter track length due to changes in image perspective or lighting, which limits the accuracy of VIO tracking.
    A new trend in feature matching now is how to combine optical flow and descriptor matching. Some existing works\cite{bang2017camera,zong2017improved,zhong2023improved} have made preliminary attempts, but they just simply combine the two algorithms together without tight fusion, which limit the efficiency and accuracy of feature matching.
\section{System Description}
\subsection{System Overview}
    Our system follows a standard visual inertial odometry framework akin to MSCKF\cite{wu2015square}. It takes inputs of images and IMU data and outputs 6DoF poses. As illustrated in \cref{fig:teaser}, the system comprises two pivotal modules: initialization and feature matching, both fundamental components in most VIO systems.
    We will provide detailed introductions to these modules. Part of initialization process is elaborated in \cref{sec:Initialization}, and part of feature matching is discussed in \cref{sec:Hybrid-Feature-Tracking}.
    
%% copy from RNIN-VIO as baseline %%
\subsection{Notation}
Here, we will define some notations and definitions that will be used consistently throughout the paper.
The world frame is denoted by $(\cdot)^w$, the camera frame by $(\cdot)^c$, and the body or IMU frame by $(\cdot)^I$. Rotation is represented using both rotation matrices, denoted by $\textbf{R}$, and Hamilton quaternions, denoted by $\textbf{q}$, with
$\otimes$ symbolizing quaternion multiplication. The gravity vector in the world frame is represented as
${\textbf{g}}^w = [0, 0, g]^T$, aligning with the z-axis of the world frame.
Lastly, $\hat{(\cdot)}$ signifies the noisy measurement or estimate of a certain quantity.

\subsection{State Definition}
  Similar to typical Extended Kalman Filter (EKF)-based VIO systems, at frame $k$, we define the system's state vector as follows:
  \begin{equation}
    {\textbf{S}_k} = [{\textbf{S}}_{I_{k-(n-1)}}, ... , {\textbf{S}}_{I_k}, {\textbf{S}}_{E_k}, {\textbf{S}}_{t}],
  \end{equation}
  for $i = k-(n-1), ...,k$, where ${S}_{I_i}$ denotes the state vector of $n$ cloned IMU poses at frame $i$. Each cloned IMU pose includes an orientation ${\textbf{q}}^w_{I_i}$ and a position   ${\textbf{p}}^w_{I_i}$. The term $\textbf{S}_{E_k}$ denotes the extra part of state on frame $k$, which consists of biases of gyroscope $\textbf{b}_{g_k}$ and accelerometer ${\textbf{b}}_{a_k}$, and also the velocity ${\textbf{v}}^w_{I_k}$ in the world frame. The term of ${\textbf{S}}_{t}$ represents time-related calibration parameters, including the IMU-camera time offset, and the rolling shutter time of the camera.

\subsection{IMU measurements} 
  IMU which including a gyroscope and an accelerometer, can measure the mobile device’s linear acceleration and angular velocity at a high frequency. However, measurements from consumer-level devices are susceptible to white gaussian noise(zero-mean gaussian noise ${\textbf{n}}_{a_k}$, ${\textbf{n}}_{g_k}$) and time-varying biases (${\textbf{b}}_{g_k}, {\textbf{b}}_{a_k}$). The gyroscope and accelerometer measurements at frame $k$, denotes as 
   ${\hat{\textbf{w}}}_{k}$ and ${\hat{\textbf{a}}}_{k}$ respectively, are given by:
  \begin{equation} 
    \begin{aligned}
      {\hat{\textbf{w}}}_{k} &= {\textbf{w}}_k + {\textbf{b}}_{g_k} + {\textbf{n}}_{g_k} \\
      {\hat{\textbf{a}}}_{k} &= {\textbf{a}}_k + {\textbf{b}}_{a_k} + {\textbf{R}}^{I_k}_{w}\,{\textbf{g}}^w + {\textbf{n}}_{a_k},
    \end{aligned}
  \end{equation}
  where ${\textbf{w}}_k$ and ${\textbf{a}}_k$ represent the ground truth states at frame $k$.
  
   % propagation
   Pre-integration is performed to integrate these measurements between consecutive frames of images. The variables are expressed in the local coordinate of frame $k$, consistent with the implementation in  ~\cite{forster2017manifold-preintergration,qin-tro-2018_VINS-Mono}: 
    \begin{equation}
      \label{eq:preintergration}
      \begin{aligned}
        \bm{\hat{\alpha}}^{I_k}_{I_{i+1}} &= {\bm{\hat{\alpha}}^{I_k}_{I_i}} + {\bm{\hat{\beta}}^{I_k}_{I_i}}\Delta t 
        + \frac{1}{2}\textbf{R}({\bm{\hat{\gamma}}^{I_k}_{I_i}})(\hat{\textbf{a}}_i - \textbf{b}_{a_k})\Delta t^2\\
        \bm{\hat{\beta}}^{I_k}_{I_{i+1}} &= {\bm{\hat{\beta}}^{I_k}_{Ii}} 
        + \textbf{R}({\bm{\hat{\gamma}}^{I_k}_{I_i}})(\hat{\textbf{a}}_i - \textbf{b}_{a_k})\Delta t \\
        \bm{\hat{\gamma}}^{I_k}_{I_{i+1}} &= {\bm{\hat{\gamma}}^{I_k}_{Ii}}Q((\hat{\textbf{w}}_i
        - \textbf{b}_{g_k})\Delta t),
      \end{aligned}
    \end{equation}
    Where $Q(\cdot)$ represents the transformation which will convert the representation of $Euler\ Angles$ to $Quaternion$. Then, the residuals of IMU measurements based on \cref{eq:preintergration} are defined as:
    \begin{equation}
    \label{IMU_residual}
      \begin{split}
      \begin{aligned}
      {\textbf{r}_I}(k) &= 
       \left[
        \begin{matrix}
          \textbf{R}^{I_k}_w({\textbf{p}_{I_{k+1}}^w} - {\textbf{p}^w_{I_k}} + \frac{1}{2}\textbf{g}^w\Delta t^2_k 
            -{\textbf{v}^w_{I_k}}\Delta t_k) -\hat{\bm{\alpha}}^{I_k}_{I_{k+1}} \\
          \textbf{R}^{I_k}_w({\textbf{v}_{I_{k+1}}^w} + \textbf{g}^w\Delta t_k - \textbf{v}^w_{I_k}) - \hat{\bm{\beta}}^{I_k}_{I_{k+1}} \\
          2[{(\textbf{q}^w_{I_k})}^{-1} \otimes \textbf{q}^w_{I_{k+1}} \otimes {(\hat{\bm{\gamma}}^{I_k}_{I_{k+1}})}^{-1}]_{xyz} \\
          \textbf{b}_{a_{k+1}} - \textbf{b}_{a_k} \\
          \textbf{b}_{g_{k+1}} - \textbf{b}_{g_k} \\
        \end{matrix}
       \right]. \\
       % &= ||{H}_{b_k}\tilde{{S}}_{b_k} + {H}_{b_{k+1}}\tilde{{S}}_{b_{k+1}} + 
       % {H}_{E_{k+1}}\tilde{{S}}_{E_{k+1}} - {r}_{u_{k+1}}||
      \end{aligned}
      \end{split}
    \end{equation}
    We also define some cost terms for optimization:
    \begin{equation}
    \label{IMU_cost}
    {C_I} = \sum_{k \in I} ||\textbf{r}_I||^2_{P^{I_k}_{I_{k+1}}},
    \end{equation}
    \begin{equation}
    \label{eq:gyro_cost}
    {C_I}(\bm{\gamma}) = \sum_{k \in I} ||\textbf{r}_I(\bm{\gamma}_k)||^2_{\textbf{P}^{I_k}_{I_{k+1}}}, 
    \end{equation}
    where $\textbf{P}^{I_k}_{I_{k+1}}$ is the covariance matrix, which is same as in ~\cite{qin-tro-2018_VINS-Mono}.  $I$ is the set where IMU measurements should be integrated. $C_I$ is the full pre-integration cost term, and $C_I(\bm{\gamma})$ is the gyroscope-related cost term aimed at constraining rotation.
\subsection{Visual measurements}
Our camera model follows the simple pinhole model, where visual measurements are the reprojection errors on image planes. 3D points are represented in the form of inverse depth. Feature $l$, which is first observed in the $i^{th}$ image, has its visual residual defined when observed again in the $j^{th}$ image as:
 \begin{equation} 
    \begin{aligned}
      \textbf{r}_V(l, j)&=\pi(\textbf{R}_w^{c_j}\textbf{X}_l+\textbf{p}_w^{c_j}) - [u^{c_j}_l,v^{c_j}_l]^T\\
      \textbf{X}_l &= \textbf{R}^w_{c_i}(d_l\cdot{\pi}^{-1}[u^{c_i}_l,v^{c_i}_l]^T) + \textbf{p}^w_{c_i},
    \end{aligned}
  \end{equation}
 where $[u^{c_i}_l,v^{c_i}_l]^T$ is the first observation of the $\textbf{X}_l$ feature that occurred in image $i$, and $[u^{c_j}_l,v^{c_j}_l]^T$ is another observation of $\textbf{X}_l$ in image $j$.  $\pi_c$ is the projection function that converts the normal coordinate to pixel coordinate with camera intrinsic parameters, and $\pi_c^{-1}$ is the back projection. $[\textbf{R}^w_{c_i},\textbf{p}^w_{c_i}]$ represents frame $i$'s camera pose in world coordinate, and $[\textbf{R}_w^{c_j}, \textbf{p}_w^{c_j}]$ represents frame $j$'s pose in camera coordinate. $\textbf{X}_l$ is the 3d coordinate of feature $l$.
 We also define the visual cost term for optimization:
 \begin{equation}
 \label{eq:cost_camera}
     C_V = \sum_{(l, j)\in V} ||{\textbf{r}_V}_{(l, j)}||^2_{\textbf{P}^{c_j}_l},
 \end{equation}
    where $V$ is the set of features that have been observed at least twice in the sliding window. $(l,j)$ denotes feature $l$ has been observed in frame $j$. ${\textbf{P}^{c_j}_l}$ is the covariance matrix for ${\textbf{r}_V}_{(l, j)}$.

\subsection{Sliding Window Filter} 
    Our multi-sensor fusion system  integrates vision and IMU measurements, aiming to minimize the cost function $C_{k}$ at each time step $k$:
  \begin{equation}
    C_{k}=C_{k-1}+C_I+C_V,
  \end{equation}
  where $C_{k-1}$ denote the cost terms related to the prior information obtained from previous time step, $C_I$, and $C_V$ denote the cost terms associated with IMU data, and visual data, respectively. These terms are consistent with those used in traditional MSCKF-based VIO systems. 
  We utilize square root inverse filter\cite{maybeck1982stochastic-square-root, wu2015square} to solve the cost function $C_{k}$ due to its computational efficiency.
 The main difference from \cite{wu2015square} is that ours does not incorporate SLAM features into the state vector, which are typically continuously updated over time. Based on our practical experimentation, we have observed that SLAM features do not significantly enhance accuracy in our system.
   
\section{State Initialization}
\label{sec:Initialization}
\begin{figure*}[h]
 \centering % avoid the use of \begin{center}...\end{center} and use \centering instead (more compact)
 \includegraphics[width=\textwidth]{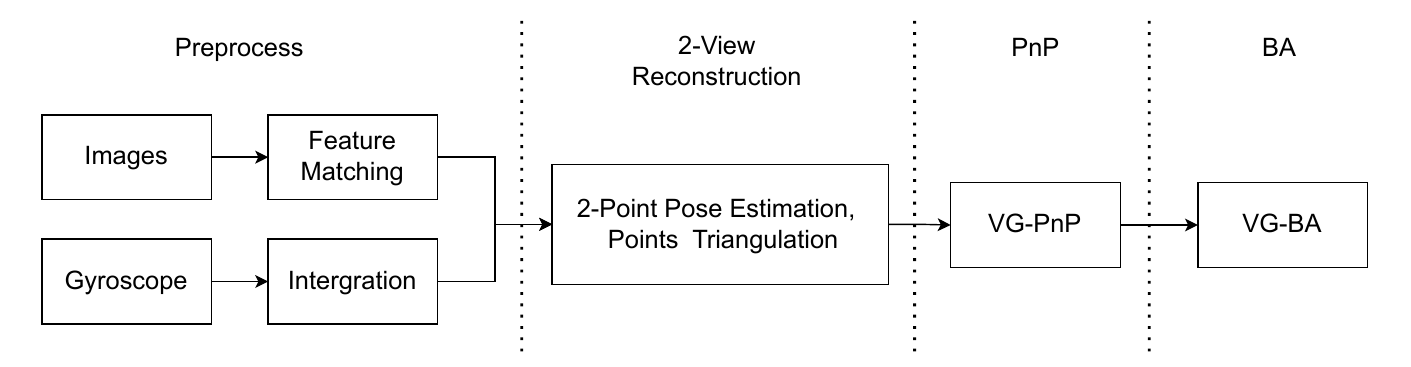}
 \caption{Pipeline of Visual Gyroscope tightly coupled SFM (VG-SFM)}
 \label{fig:VG-SFM}
\end{figure*}

\begin{figure}[h]
 \centering % avoid the use of \begin{center}...\end{center} and use \centering instead (more compact)
 \includegraphics[width=\columnwidth]{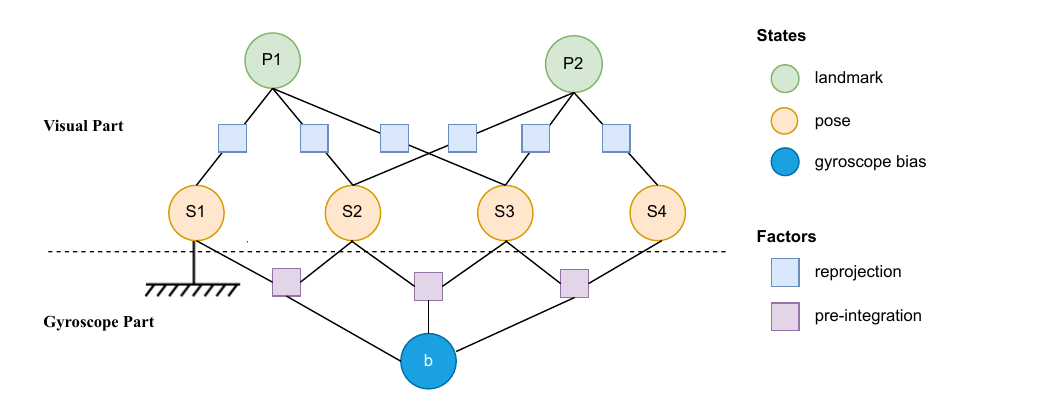}
 \caption{Factor Graph of VG-BA}
 \label{fig:VG-BA-F}
\end{figure}

A high-quality initialization can significantly accelerate the convergence speed of the filter. We employ two initialization methods depending on the motion states, namely static and motion.
To determine the current state, we consider the average displacement of sparse features and the standard deviation of acceleration and angular velocity. If both the average displacement and standard deviation are below specific thresholds, we perform static initialization similar to OpenVINS \cite{geneva2020openvins}; otherwise, we proceed with motion initialization.

\subsection{Motion Initialization}
We design a new VI initialization method to robustly handle the fast VIO motion initialization problem. In this case, we need to solve the problem based on very few observations, typically using only 4 image frames and related IMU data.
Given the limited number of image observations, traditional SfM solutions become highly fragile and prone to erroneous or failed solutions.

We revisit the problem of visual IMU initialization. There are a large number of parameters that need to be estimated, and visual observations alone are not sufficient. Although the IMU is noisy, the gyroscope can often provide relatively accurate initial orientation, which is irrelevant to the scene. This is also helpful for visual SfM. However, the state of the accelerometer is very difficult to estimate. So we need to wait for some parameters to be stable, and then initialize the accelerometer-related parameters, such as scale, velocity and gravity. Based on these studies, we designed a new pipeline of VI initialization. As shown in \cref{fig:teaser}, the core idea is to first tightly couple with the gyroscope and then loosely couple with the accelerometer, and finally optimize all parameters together. 
The pipeline consists of VG-SfM, VA-Align and VI-BA.  Especially for VG-SfM, rotation from the gyroscope's integration is tightly used in the whole process of SfM, which significantly increases the robustness and accuracy of SfM.
\subsection{VG-SfM}
VG-SfM is a tightly coupled method that integrates visual and gyroscope data. In scenarios with short time intervals and minimal parallax, visual measurements may lack stability. Conversely, the gyroscope provides high-precision rotation information, even in the absence of correct bias. Leveraging this, we utilize the rotation information obtained from gyroscope integration as a vital prior and constraint for visual SfM, resulting in a more stable SfM solution. Refer to \cref{fig:VG-SFM} for an illustration.
\begin{enumerate}
    \item \textbf{Preprocess}: We use visual and inertial data for system initialization. Upon capturing images, we initially perform feature tracking with \cite{Lucas_Kanade_1981_KLT} . Subsequently, we integrate the angular velocity data from the gyroscope \cite{forster2017manifold-preintergration}.  Before two-view reconstruction, we select two frames with the maximal parallax as keyframes from the initial frames.
    \item \textbf{Two View Reconstruction}: Traditional visual SfM requires 5-point correspondences \cite{fivepoint2004nister} to estimate relative pose, which is accurate in most case but can be unreliable in scenarios with limited parallax. However, in the context of VIO, we can obtain initial rotation information from the gyroscope. Despite potential noise and long-term drift in gyroscope data, it remains accurate and robust over short time intervals. Therefore, we opt to directly utilize gyroscope rotation instead of relying solely on visual cues. With the rotation known, the translation parameters to be estimated reduce to just 2 DoF (excluding scale), rendering the problem linear and amenable to robust solutions. This simplification facilitates the identification of inliers and determination of optimal parameters using \textbf{2-Point-RANSAC} \cite{Kneip_Chli_Siegwart_2010}. Subsequently, initial pose computation enables triangulation of 3D points, facilitating solution of other frames based on these points.
    \item \textbf{VG-PnP}: Given a sufficient number of 3D points and corresponding 2D features, we employ Perspective-n-Point (PnP) to determine the pose of each frame. Also, we utilize gyroscope's rotation measurements to improve the robustness of visual PnP, a method we term VG-PnP, which tightly integrates visual and gyroscope data. The problem can be formulated as follows:
    \begin{equation}\label{eq:vg_pnp}
        {\arg\min_{\textbf{R}_k,\textbf{t}_k}} ( {C_I}_{k}(\bm{\gamma}) + {C_V}_{k}),
    \end{equation}
where ${C_I}_{k}(\bm{\gamma})$ represents the pre-integration term of IMU for frame $k$, utilizing gyroscope measurements to constrain rotation. ${C_V}_k$ is the visual term from image frame $k$, using 3D-2D measurements to constrain 6DoF pose. For frames within the sliding window excluding the two initialized frames, we consider the minor camera motion during the initialization process. We use the position of the initialized neighboring frame as the initial position and accumulate the pre-integrated rotation from the initialized neighboring frame to the current frame as the initial rotation. Subsequently, we apply the Levenberg-Marquardt method to minimize \cref{eq:vg_pnp}, thereby obtaining the pose for each uninitialized frame.

        \item \textbf{VG-BA}: Upon calculating the initial values of 3D points and camera poses, we proceed to perform VG-BA, , a bundle adjustment tightly coupling visual and gyroscope data. The initial value of gyroscope's bias is set to zero. VG-BA comprises two components, where both residuals are combined and optimized to minimize the total cost, formulated as:
          \begin{equation}
      \label{VG-BA}
      \begin{aligned}
{\arg\min_{\textbf{X}_l,\textbf{R}_k,\textbf{t}_k,\textbf{b}_g}} ( C_I(\bm{\gamma}) + C_V),
      \end{aligned}
    \end{equation}
    where $C_I(\gamma)$ is defined in \cref{eq:gyro_cost}, and $C_V$ is defined in \cref{eq:cost_camera}. $\textbf{X}_l, \textbf{R}_k, \textbf{t}_k, \textbf{b}_g$ represent the states to be estimated. $\textbf{X}_l$ denotes the $l^{th}$ 3D point, $\textbf{R}_k$ and $\textbf{t}_k$ denote the pose of frame $k$, and $\textbf{b}_g$ represents the bias of gyroscope. It is important to note that the pose of the first frame needs to be fixed. Further details of the factor graph can be found in \cref{fig:VG-BA-F}.
    After VG-SfM, we obtain the pose of each frame. In order to align them with IMU data, we transform them into body frame. 
 \end{enumerate}
    \subsection{VA-Align} Following a similar approach to \cite{qin-tro-2018_VINS-Mono}, we initially define the initial state vector for visual inertial alignment as:
    
  \begin{equation} \label{eq:init_state}
    \textbf{X}_{init} = [{\textbf{v}}_{I_0}, ... , {\textbf{v}}_{I_n}, s, \textbf{g}^{c_0}],
  \end{equation}
  where ${\textbf{v}}_{I_k}$ represents the velocity in the body frame $k$, $s$ is the scale factor between visual SfM and IMU, and $\textbf{g}^{c_0}$ is the gravity vector in frame $c_0$.  
    By defining the state as in \cref{eq:init_state}, we can rewrite the \cref{eq:preintergration} as:
        \begin{equation}
      \label{eq:VI-Align-integration}
      \begin{aligned}
        \bm{\alpha}^{I_k}_{I_{k+1}} &= \textbf{R}^{I_k}_{c_0}(s(\textbf{p}^{c_0}_{I_{k+1}} - \textbf{p}^{c_0}_{I_{k}})) + \frac{1}{2}\bm{g}^{c_0}\Delta t_k^2 - \textbf{R}^{c_0}_{I_k}\textbf{v}_{I_k}\Delta t_k)\\
        \bm{\beta}^{I_k}_{I_{k+1}} &= {\textbf{R}^{I_k}_{c_0}}({\textbf{R}^{c_0}_{I_{k+1}}}\textbf{v}_{I_{k+1}} 
        + \textbf{g}^{c_0}{\Delta}{t_k} 
        - {\textbf{R}^{c_0}_{I_k}}v_{I_k}).
      \end{aligned}
    \end{equation}
   We combine \cref{eq:VI-Align-integration} and \cref{eq:preintergration} and solve for the initial state $X_{init}$, following the methodology outlined in \cite{qin-tro-2018_VINS-Mono}.
   %TODO BA function
    \subsection{VI-BA} Once all the initial states are estimated, we perform an overall bundle adjustment to further improve the accuracy. Unlike traditional VI-BA methods with fixed weights, we have developed a scheme to adjust the weights of the visual and IMU term based on disparity. It has been observed that when the disparity is small, the residuals of $C_{total}$  are primarily influenced by $C_I$ alone, while the the degrees of freedom of the 4KF IMU are excessively high. As a result, the optimization of VI-BA tends to minimize IMU errors rather than overall errors. Below are the revised formulas:
    \begin{equation}
    \begin{aligned}
        C_{total} &= w(P) *C_V + C_I\\
        w(P) &= \frac{w_{max}}{1+e^{(P-P_{min})}} + w_{min},
    \end{aligned}
    \end{equation}
    where $w(P)$ represents a sigmoid-like weighting function used to to balance the weights of the visual and IMU term. $P$ denotes the average parallax of the two frames with the largest parallax in SfM. $w_{max}, w_{min}$ are the maximal (approximative maximal) and minimal values of the weighting function, respectively. In this work, we set $(w_{max}, w_{min})$ to $(e^4, 1)$. $P_{min}$ is the threshold representing the minimum parallax range within which BA can work effectively. In this work, we set it to 20 pixels.

    During the bundle adjustment process, we fix the position and yaw of the first frame because these 4 degrees of freedom are unobservable.
       
\section{Feature Matching}
\label{sec:Hybrid-Feature-Tracking}
\begin{figure}[h]
    \centering
    \includegraphics[width=1\linewidth]{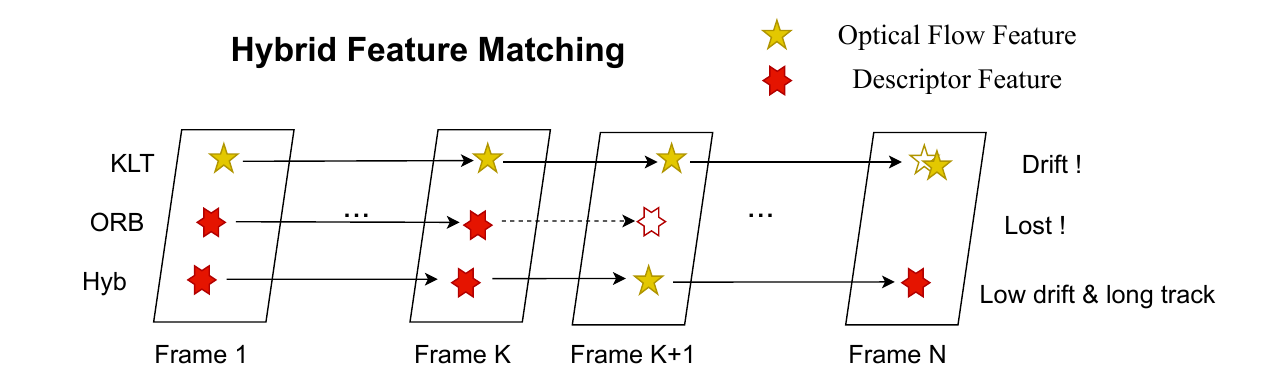}
    \caption{Hybrid Feature Matching Approach}
    \label{fig:Feature—Tracking}
\end{figure}

Traditional VIO feature matching methods are typically classified into two categories: optical flow-based \cite{qin-tro-2018_VINS-Mono,geneva2020openvins} and descriptor matching-based \cite{leutenegger-ijrr-2015-OKVIS,campos2021orb-slam3}. Optical flow-based methods, such as KLT\cite{Lucas_Kanade_1981_KLT,opencv_library}, often exhibit continuous drift, which limits VIO accuracy. Conversely, descriptor-based methods may suffer from a lack of long-term tracking capability. To address these limitations, the community has recently proposed some joint tracking solutions, such as \cite{zong2017improved,zhong2023improved,bang2017camera}, etc. However, some\cite{zong2017improved,zhong2023improved} of these methods simply combine two features by just using feature extraction points of ORB for optical flow tracking, or like\cite{bang2017camera}, just running two matching methods separately and then selecting the better one. These algorithms could not fully leverage the advantages of both methods, which limits the upper bound of feature matching. So we introduce a novel hybrid feature matching scheme which tightly integrates optical flow and descriptor methods. As illustrated in \cref{fig:Feature—Tracking}, features can be tracked using both optical flow and descriptor, aiming to achieve a balanced trade-off between track length and accuracy.

%Traditional methods for feature matching in Visual-Inertial Odometry (VIO) can be broadly categorized into two types: optical flow-based approaches and descriptor matching-based methods. Optical flow-based methods, such as KLT (Lucas-Kanade Tracker), are known for their tendency to suffer from continuous drift, thus limiting the overall accuracy of VIO systems (Qin et al., 2018; Geneva et al., 2020). On the other hand, descriptor-based methods, like ORB-SLAM3 (Leutenegger et al., 2015; Campos et al., 2021), may face challenges in maintaining long-term tracking capabilities.

%To mitigate these shortcomings, recent efforts in the research community have explored hybrid solutions that combine both optical flow and descriptor matching techniques (Zong et al., 2017; Zhong et al., 2023; Bang et al., 2017). However, some of these hybrid methods simply concatenate features or independently apply different matching algorithms without fully leveraging the strengths of each method. This approach often limits the potential upper bound of feature matching performance.

%In response, this paper introduces a novel hybrid feature matching scheme that integrates optical flow and descriptor methods more closely. This integrated approach, illustrated in Figure 1, aims to achieve a balanced trade-off between track length and accuracy in feature tracking.

Specifically, as outlined in \cref{alg:feature-matching}, for each frame of image, we first perform feature detection and description. Subsequently, during the feature matching process, tracks with established triangulation can be projected onto the current frame as priors for matching. The initial pose of the current frame is derived from IMU propagation. Leveraging these priors, descriptor-based matching is performed, followed by a ratio test to validate good matches. This is similar to ORB-SLAM\cite{murartal-tro-2015-ORB-SLAM}, with the distinction that we only utilize feature tracks within the current sliding window.

For feature tracks unable to be triangulated or have failed matches through projection, they transition to the second stage of matching, utilizing either 2D priors or optical flow priors. All these features will first undergo optical flow \cite{Lucas_Kanade_1981_KLT} from the previous frame to the current frame. These matches serve as priors for descriptor matching. Here we use ORB features for matching. With optical flow priors, ORB matching only requires KNN matching within a very small window (set to a radius of 10 pixels) and undergoes a ratio test (threshold set to 0.7). ORB points that fail to match still utilize the optical flow matching result for the current frame, with the descriptor retained, thus preventing interruptions in feature tracking due to descriptor matching failures and ensuring longer track lengths. If the number of tracks in the current frame falls below a certain threshold (set to 150), new tracks will be generated, awaiting subsequent feature matching.

It is important to note that while our method employs ORB as an example, it can be substituted with any descriptor-based method.
In \cref{sec:Experimental_Results}, we quantified the effectiveness of this matching strategy on public datasets, including matching accuracy indicators and VIO trajectory accuracy indicators.

\begin{algorithm}[h]
\DontPrintSemicolon
\SetKwInput{KwInput}{Input}                % Set the Input
\SetKwInput{KwOutput}{Output}              % Set the Output
\KwInput{Images, 3D Points, Poses}
\KwOutput{Matched Features}

\SetKwFunction{FMain}{}    % 先定义 \FMain
\SetKwProg{Fn}{Function}{:}{\KwRet}        % 定义函数风格
\SetKwProg{FnMain}{HybMatching}{:}{\KwRet} % 定义主函数风格
\FnMain{\FMain}{
    
    \tcc{Max 150 kpts per-frame, pyramid level 1}
    
    ORBFeatureDetect();
    
    ComputeORBDescriptor();
    
    \tcc{Descriptor matching by prior}
    
    \For{feature track $f_k$}{
        \If{$f_k$ IsTriangulated}{
            \tcc{Similar with \cite{murartal-tro-2015-ORB-SLAM}}  
            MatchWith3DProjection(); 
        }
        \Else{
            MatchWith2DPrior();
        }
    }
    Ransac(); \tcc{Remove outliers}
    \If{count(track) < 150}{
        AddNewTracks();
    }
    \KwRet{all\_matches} % 明确返回值
}
\Fn{\FMain MatchWith2DPrior{}}{
    
    \tcc{OpticalFlow as prior, from previous frame to current}
    OpticalFlow();
    
    \tcc{Search in a 10x10 pixels window by OpticalFlow's prior}
    SearchInArea(10);
    
    RatioTest(0.7);
    
    \tcc{When match failed, add OpticalFlow's result as supplement}
    AddFlowToMatches();
    
    Ransac();
}
\caption{Hybrid Feature Matching}
\label{alg:feature-matching}
\end{algorithm}   
\section{Experimental Results}
\label{sec:Experimental_Results}
We conducted a series of experiments to evaluate our proposed XR-VIO system. For initialization evaluation, we partitioned the dataset into numerous small fragments. Subsequently, VI initialization was executed fragment by fragment, and the metric error was computed for each fragment. Additionally, we run and evaluate overall trajectory's accuracy ont both MAV and phone datasets. We conducted qualitative comparisons and analyses of our method against other SOTA algorithms.

Our method and other SOTA systems were evaluated using the following public datasets:
\begin{itemize}
    \item \textbf{EuRoC Dataset}~\cite{Burri25012016-EuRoC}: This dataset features high-quality recordings gathered by a micro aerial vehicle (MAV), which is equipped with a stereo camera and an IMU. They are synchronized and pre-calibrated both in spatial and temporal aspects. It covers a range of indoor environments and includes ground truth states (VICON) to facilitate the evaluation of trajectory accuracy.
    \item \textbf{ZJU-Sensetime Dataset} \cite{jinyu2019survey}: This dataset is collected by handheld mobile phones. It contains pre-calibrated images and IMU data, with the ground truth trajectory (VICON). The dataset covers a variety of scenes and motion types, including fast motion, occlusion, rotation, straight lines, and more.
\end{itemize}

The experiments were performed on a desktop PC with Ubuntu 20.04 LTS, featuring with an $\text{Intel}^{\circledR }$ Core™ i7-7700K CPU @ 4.20GHz × 8 and 32 GB memory. 

\subsection{Baseline Methods}
For the comparison of initialization methods, we utilize Closed-form \cite{martinelli2014closed}, VINS-Mono\cite{qin-tro-2018_VINS-Mono}, Inertial-only \cite{campos2020inertial} and the latest DRT\cite{Rotation-Translation-Decoupled} as baselines. 
In evaluating the overall trajectory accuracy, we employ OKVIS \cite{leutenegger-ijrr-2015-OKVIS}, VINS-Mono\cite{qin-tro-2018_VINS-Mono}, VINS-Fusion\cite{qin2019a_VINS_Fusion_Local}, OpenVINS\cite{geneva2020openvins} and HybVIO\cite{hybvio} as baselines. These methods are widely recognized and offer state-of-the-art accuracy. All selected baselines are open source, and we employ their default code and configurations for comparison purposes.

\subsection{Metrics Definition}
For the evaluation of initialization, we employ three metrics: absolute trajectory error (ATE)\cite{Zhang18iros-Quantitative-Trajectory-Evaluation} , scale error and gravity error. To facilitate the evaluation of trajectory metrics, we utilize evo \cite{grupp2017evo}, an open-source Python tool. We define the three metrics as follows:
\begin{equation}
      \label{ATE}
      \begin{aligned}
{ATE}_{p} = (\frac{1}{n}\sum_{i=0}^{n-1}|\textbf{p}_i - \hat{\textbf{p}_i}|^2)^{\frac{1}{2}},
      \end{aligned}
\end{equation}
    where $\textbf{p}_i$ is the ground truth value of position of frame $i$, and $\hat{\textbf{p}_i}$ is the estimation value of it. $ATE_{p}$ is the root mean square error (RMSE) of a trajectory.
    \begin{equation}
      \label{ScaleError}
      \begin{aligned}
S_{err} &= \frac{1}{m}\sum_{k=0}^{m-1}|\hat{s_k}^{\prime}-1|*100\%\\
    \hat{s_k}^{\prime} &= 
    \begin{cases}
    \hat{s_k}& \hat{s_k}\leq1\\
        1/\hat{s_k}& \hat{s_k}>1
    \end{cases},
      \end{aligned}
\end{equation}

where $\hat{s_k}$  is the estimated scale of the $k^{th}$ trajectory and $\hat{s_k}^{\prime}$ is the normalized scale. $S_{err}$ is the mean error of all trajectories.
\begin{equation}
      \label{GravityError}
      \begin{aligned}
G_{err} = (\frac{1}{n}\sum_{i=0}^{n-1}|
\frac{180^{\circ}}{\pi}\cdot
\arccos({\textbf{g}_i} \cdot \hat{\textbf{g}_i})|^2)^\frac{1}{2},
      \end{aligned}
\end{equation}
where $\textbf{g}_i$ is the true gravity direction of frame $i$, and $\hat{\textbf{g}_i}$ is the estimated value. $G_{err}$ is the mean error in degrees for each frame in a trajectory.

\subsection{Initialization Evaluation}
For the initialization evaluation, following a methodology similar to \cite{Mono-Depth-VI-Init-2022}, we divide each sequence into small fragments. In the case of the 4-keyframe (4KF) / 5-keyframe (5KF) test scenario, each fragment consists of image frames taken at intervals of 0.6 / 0.8 seconds along with corresponding IMU data, totaling 2293 / 1271 fragments. Keyframes are uniformly selected at intervals of 0.1 seconds for all methods. We conduct comparative experiments in two configurations: 4KF (total time is \textit{0.3} seconds) and 5KF (total time is \textit{0.4} seconds). Among them, 5KF is a common initialization configuration in most other works, while 4KF is the theoretical minimum initialization frame configuration. We aim to challenge the limits of VI initialization.

We compare two loosely-coupled initialization methods, which include VINS-Mono initialization (referred to as VINS-Mono) \cite{qin-tro-2018_VINS-Mono} and the Inertial-only method\cite{campos2020inertial}, integrated into ORB-SLAM3 \cite{campos2021orb-slam3} (referred to as Inertial-only). Additionally, for the tightly-coupled closed-form initialization method (referred to as Closed-form) \cite{martinelli2014closed}, we utilize code from the open-source SLAM OpenVINS \cite{geneva2020openvins} for comparison. To assess gyro tightly-coupled and accelerometer loosely-coupled initialization methods, we employ DRT \cite{Rotation-Translation-Decoupled}. All mentioned initialization methods, including our own, are executed using the same sample data and configuration.

\begin{table*}[t]

    \caption{Initialization Evaluation on EuRoC. Bold font indicates the best result. We compare these methods under 2 different configuration: 4KF and 5KF, both including scale, ATE, gravity, and success rate.}
    \centering
    \begin{tabular}{c|cccc|cccc}
    \toprule
 & \multicolumn{4}{c|}{4KF} &\multicolumn{4}{c}{5KF}\\
         \midrule
         &  Scale(\%)$\downarrow$&  ATE(m)$\downarrow$&  Gravity(°)$\downarrow$&Success(\%)$\uparrow$& Scale(\%)$\downarrow$& ATE(m)$\downarrow$& Gravity(°)$\downarrow$& Success(\%)$\uparrow$\\
         \midrule
         Closed-form\cite{genevaopenvins}& 29.95 & 0.033 & 2.76 & 2.17 & 25.39 & 0.031 & 2.21 & 12.86 \\
         Inertial-only\cite{campos2020inertial}&  48.84&  0.063&  10.48&57.07 & 39.09& 0.058& 12.61&64.01\\
         Vins-Mono\cite{qin-tro-2018_VINS-Mono}&  32.64&  0.034&  2.74&47.34 & 28.75& 0.037& 2.36&62.04\\
         DRT-l\cite{Rotation-Translation-Decoupled}&  41.79&  0.041&  3.50&76.94 & 34.57& 0.039& 2.64&85.79\\
         DRT-t\cite{Rotation-Translation-Decoupled}&  61.54&  0.059&  3.97&\textbf{86.56} & 50.46& 0.059& 3.37& 86.24\\
         XR-VIO&  \textbf{26.88}&  \textbf{0.026}&  \textbf{2.26}&83.98 & \textbf{22.71}& \textbf{0.027}& \textbf{1.99}&\textbf{87.15}\\
         \bottomrule
    \end{tabular}
    \label{tab:Init_Compare}
\end{table*}

As shown in \cref{tab:Init_Compare}, in both 4KF and 5KF configurations, our method achieves the highest accuracy in scale, ATE and gravity direction. However, it's essential to note that success here merely indicates the capability of the initialization module to process current fragment and produce initialization poses. Success does not necessarily imply that the initialization results meet a certain threshold, nor does it imply that the initialization poses are qualified for VIO tracking. Actually, DRT-l and DRT-t demonstrate high success rates due to their decoupling of rotation and translation. By employing Kneip's\cite{Kneip_Lynen_2013-rotation} and Cai's\cite{Cai_2021-pose-only} methods, DRT bypasses the SfM solution process, significantly enhancing the initialization robustness even in some challenging scenarios. However, based on our experiences, we assert that SfM and bundle adjustment remain the optimal choice. This scheme can fully utilize visual and IMU measurements. Given that VI initialization poses a highly non-linear problem, and DRT's pose-only scheme prevents optimization with bundle adjustment, potentially leading to decreased accuracy.

% \begin{figure*}[h]
%     \centering
%     \includegraphics[width=1\linewidth]{pictures/Trajectory-Visual-Init.png}
%     \caption{Visualization of scale error on V2-01 (EuRoC). Fragments of poses are color-coded according to the magnitude of scale error for each initialization window in the dataset. Darker colors represent greater error, lighter colors indicate lower error, and black denotes failed initializations.}
%     \label{fig:Trajectory-Visual-Init}
% \end{figure*}

\begin{figure*}[h]
    \centering
    \begin{minipage}[t]{0.183\textwidth}
        \centering
        \includegraphics[width=\textwidth]{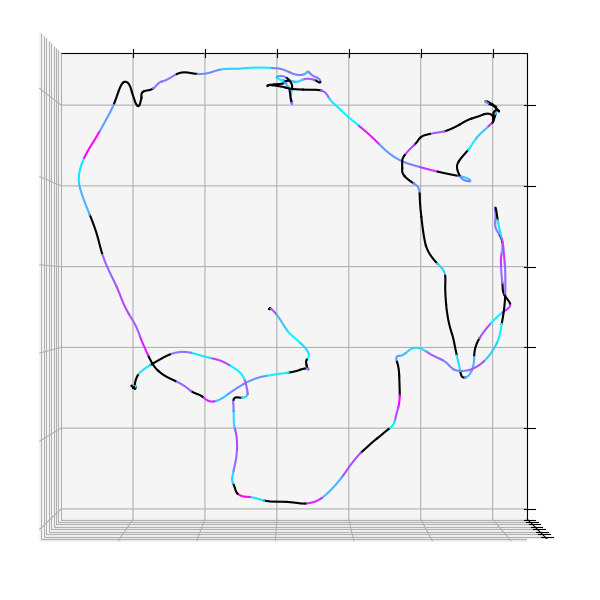} % 图片1
        \parbox{\textwidth}{\centering Inertial-only}
    \end{minipage}%
    \begin{minipage}[t]{0.183\textwidth}
        \centering
        \includegraphics[width=\textwidth]{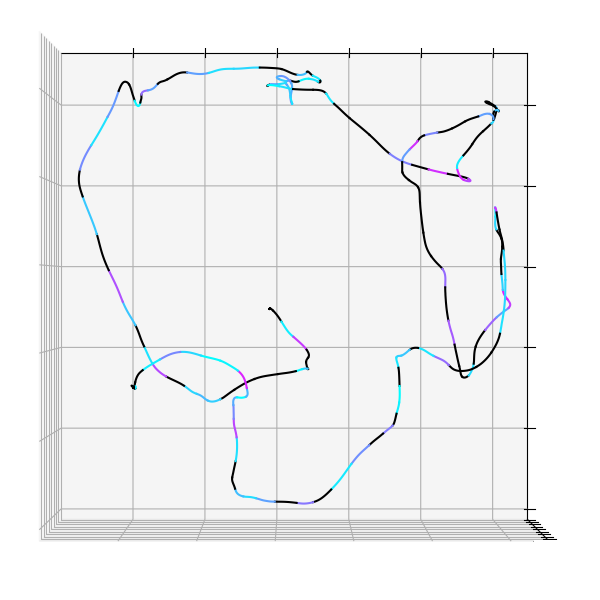} % 图片2
        \parbox{\textwidth}{\centering Vins-Mono}
    \end{minipage}%
    \begin{minipage}[t]{0.183\textwidth}
        \centering
        \includegraphics[width=\textwidth]{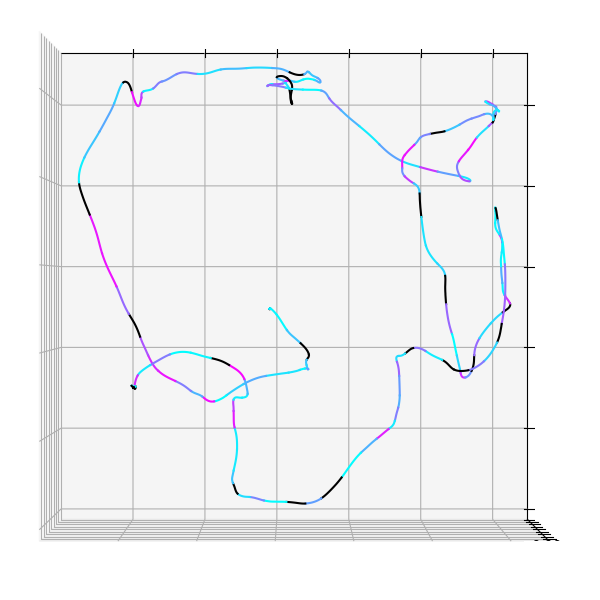} % 图片3
        \parbox{\textwidth}{\centering DRT-l}
    \end{minipage}%
    \begin{minipage}[t]{0.183\textwidth}
        \centering
        \includegraphics[width=\textwidth]{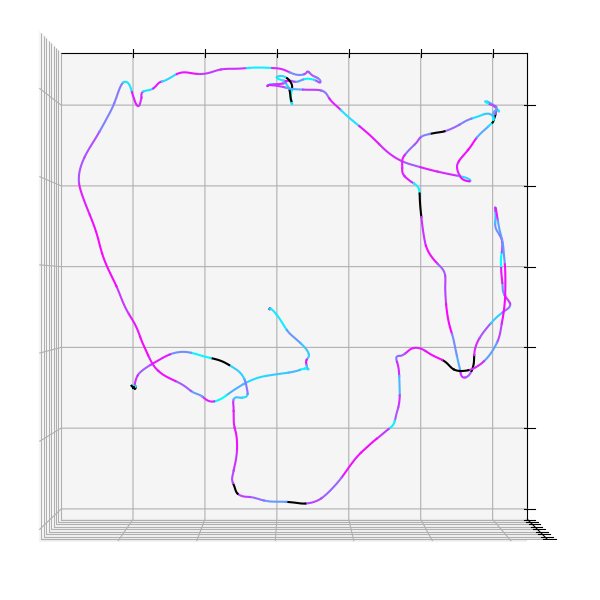} % 图片4
        \parbox{\textwidth}{\centering DRT-t}
    \end{minipage}%
    \begin{minipage}[t]{0.183\textwidth}
        \centering
        \includegraphics[width=\textwidth]{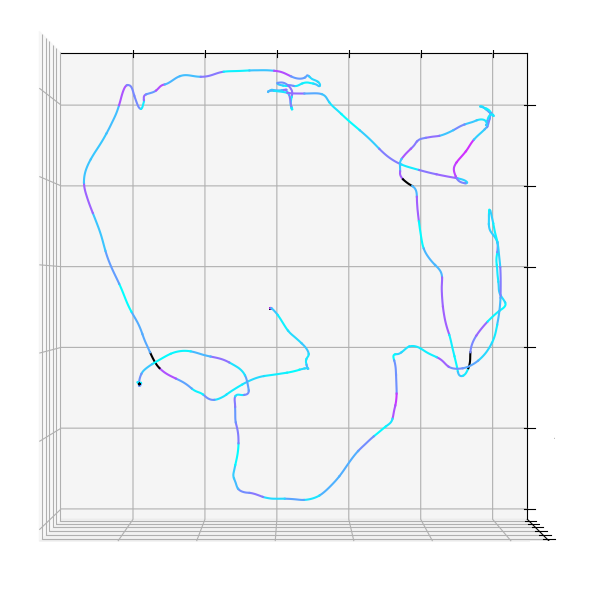} % 图片5
        \parbox{\textwidth}{\centering XR-VIO}
    \end{minipage}%
    \begin{minipage}[t]{0.055\textwidth}
        \centering
        \begin{minipage}[t]{1\textwidth}
            \centering
            \includegraphics[width=\textwidth]{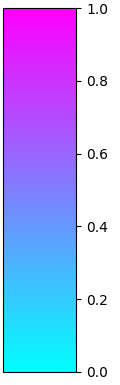} % 图片6上半部分
            \parbox{\textwidth}{\centering Scale Error }
        \end{minipage}
    \end{minipage}%
    \caption{Visualization of scale error on V2-01 (EuRoC). Fragments of poses are color-coded according to the magnitude of scale error for each initialization window in the dataset. Darker colors represent greater error, lighter colors indicate lower error, and black denotes failed initializations.}
    \label{fig:Trajectory-Visual-Init}
\end{figure*}

In \cref{fig:Trajectory-Visual-Init}, we conducted a qualitative comparison of various methods. Similar to \cite{Mono-Depth-VI-Init-2022}, we visualized the V2-01 sequence from the EuRoC dataset, color-coding the scale error for each initialization fragment. Our approach demonstrates superior performance over existing methods in terms of both initialization success rate and accuracy, showcasing the effectiveness of our initialization scheme across a range of motion modes and providing a more user-friendly VIO initialization experience.

\begin{figure}[!h]
    \centering
    \includegraphics[width=1\linewidth,height=0.032\linewidth]{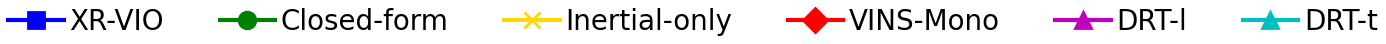}
    4KF
    \centering
    \includegraphics[width=1\linewidth,height=0.3\linewidth]{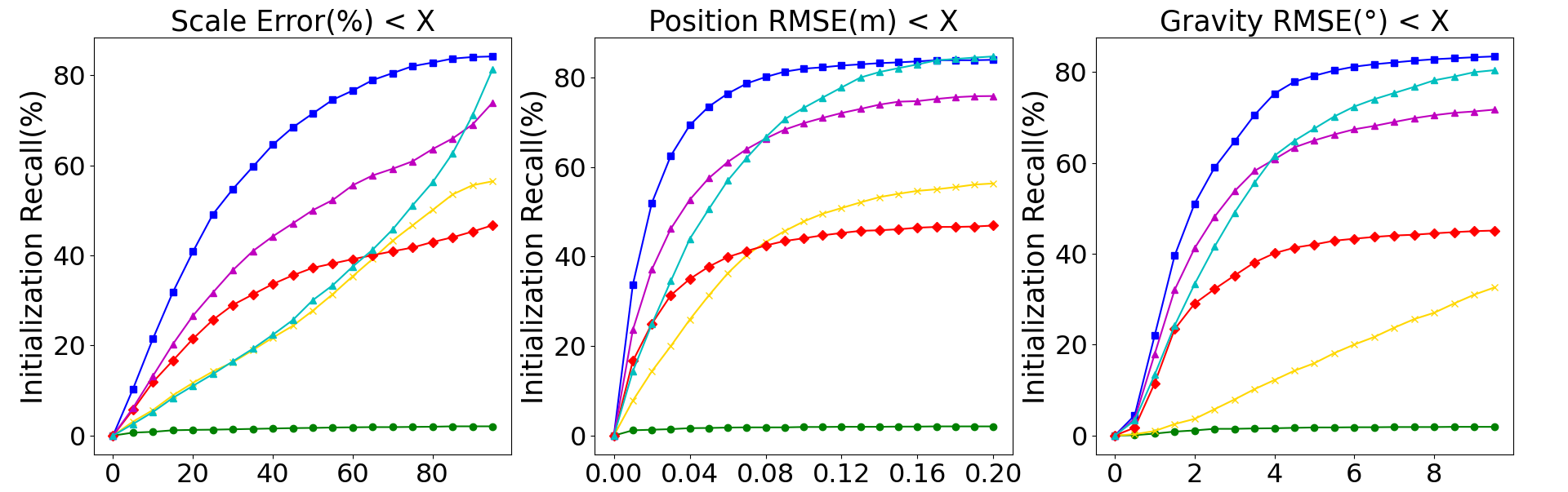}
    5KF
    \includegraphics[width=1\linewidth,height=0.3\linewidth]{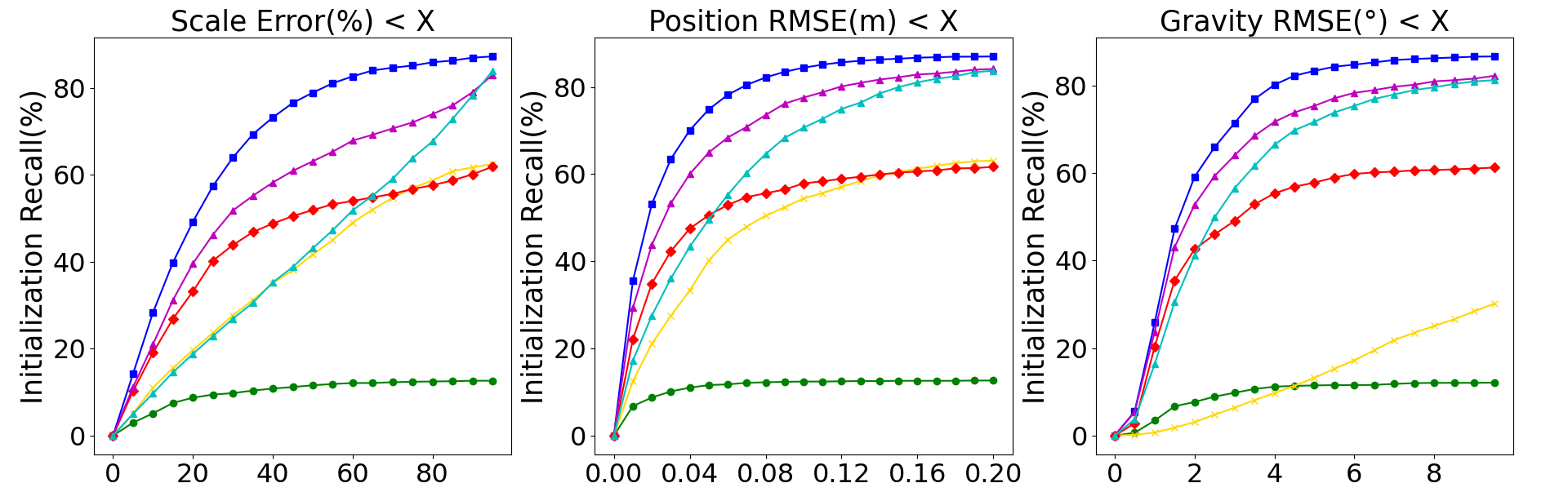}
    \caption{Cumulative distribution of initialization with different keyframes: 4KF and 5KF. Scale error, ATE and gravity RMSE are shown in 3 columns. }
    \label{fig:Init_Error_CDF}
\end{figure}

It is important to note that the accuracy values here are averages of all successful cases. A high success rate may lower the average accuracy. Therefore, to provide a more comprehensive evaluation, we further plotted the cumulative distribution function (CDF) \cite{Schinazi2022-CDF} in \cref{fig:Init_Error_CDF}. From \cref{fig:Init_Error_CDF}, it is evident that although DRT achieves higher success rate, the accuracy of many successful cases is very low. In contrast, our method achieves higher accuracy at each equivalent success rate. The closed-form method exhibits a particularly low success rate, primarily due to significant algorithmic instability in scenarios with very few keyframes.

We also analyzed the time consumption of initializing each module, as shown in \cref{tab:time_statistics}. The primary time-consuming modules include 2-View Reconstruction, VG-BA and VI-BA. Among these, 2-View Reconstruction primarily consumes time in the triangulation of points, while VG-BA and VI-BA  are mainly consumed in nonlinear optimization. The time-consuming statistics are calculated under the 4KF configuration.  On average, each frame takes about 13 ms to initialize, meeting real-time requirements.

%% 3 x 2, (4KF, 5KF) x (ATE, scale, gravity) %

\begin{table}[h]
    \caption{Time Statistics of Each Submodule in VIO}
    \centering
    \begin{tabular}{cp{0.35\columnwidth}cc}
    \toprule
    & \textbf{Submodule} & \multicolumn{2}{c}{\textbf{Time (ms)}} \\
    \midrule
    \multirow{5}{*}{Initialization} & 2-View Reconstruction & \multicolumn{2}{c}{24.24} \\
    & VG-BA & \multicolumn{2}{c}{12.12} \\
    & VA-Align & \multicolumn{2}{c}{0.09} \\
    & VI-BA & \multicolumn{2}{c}{14.86} \\
    %\cdashline{2-4}
    % \cline{2-4}
    & \textbf{Total} & \multicolumn{2}{c}{\textbf{55.45}} \\
    \midrule
    % &&KLT& Hyb\\
    % \midrule
    \multirow{5}{*}{Tracking} &&KLT& Hyb\\ & Feature Detection&4.2& 3.9\\
    & Feature Matching&4.1& 7.9\\
    & Others& 5.7&4.4\\
    %\cdashline{2-4}
    % \cline{2-4}
    & \textbf{Total}&\textbf{14.0}& \textbf{16.2}\\
    \bottomrule
    \end{tabular}
    \label{tab:time_statistics}
\end{table}

 \subsection{Hybrid Matching Evaluation}

% \begin{figure} [h]
%     \centering
%     \includegraphics[width=1\linewidth]{pictures/Epipolar-Error.png}
%     \caption{Epipolar error statistics on all data from EuRoC. The x-axis represents track length, and the y-axis indicates the median value of the epipolar error corresponding to the track length.}
%     \label{fig:Epipolar_Error}
% \end{figure}

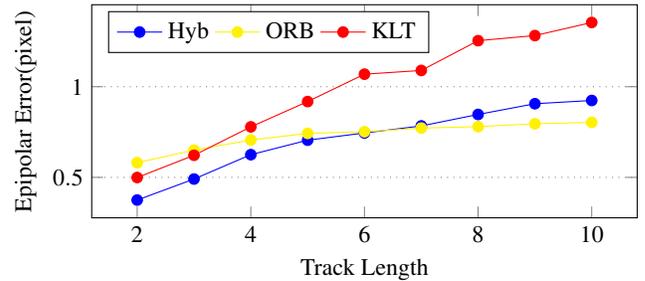
\begin{figure}[h]
    \centering
    \begin{tikzpicture}
        \begin{axis}[
            xlabel={Track Length},
            ylabel={Epipolar Error(pixel)},
            legend pos=north west, % 图例位置
            legend columns=3, % 设置图例的列数为3
            grid=both, % 显示网格
            major grid style={dotted,gray}, % 主要网格线样式
            minor grid style={dotted,gray}, % 次要网格线样式
            xmajorgrids=false, % 不显示垂直主要网格线
            xminorgrids=false, % 不显示垂直次要网格线
            width=\linewidth, % 宽度设置为12厘米
            height=0.5\linewidth, % 高度设置为4厘米
            ]
            \addplot[color=blue,mark=*] coordinates {
                (2,0.374883)
                (3,0.490805)
                (4,0.625229)
                (5,0.705148)
                (6,0.744951)
                (7,0.783779)
                (8,0.84706)
                (9,0.906202)
                (10,0.924143)
                % (11,0.989535)
                % (12,0.973194)
                % (13,1.139)
                % (14,1.08904)
                % (15,1.14394)
                % (16,1.21185)
                % (17,1.10777)
                % (18,1.4823)
                % (19,1.19149)
                % (20,1.44675)
                % (21,1.43161)
                % (22,1.44882)
                % (23,1.50112)
                % (24,1.52332)
                % (25,1.5785)
            };
            \addlegendentry{Hyb}
            \addplot[color=yellow,mark=*] coordinates {
                (2,0.581117)
                (3,0.650189)
                (4,0.706101)
                (5,0.7426)
                (6,0.751624)
                (7,0.771983)
                (8,0.779857)
                (9,0.79575)
                (10,0.803329)
                % (11,0.808082)
                % (12,0.803262)
                % (13,0.79011)
                % (14,0.798317)
                % (15,0.779228)
                % (16,0.836509)
                % (17,1.0089)
                % (18,0.964484)
                % (19,0.840002)
                % (20,0.889293)
                % (21,0.769063)
                % (22,1.06872)
                % (23,0.855994)
                % (24,1.00139)
                % (25,1.0281)
            };
            \addlegendentry{ORB}
            \addplot[color=red,mark=*] coordinates {
                (2,0.499181)
                (3,0.62185)
                (4,0.778618)
                (5,0.918091)
                (6,1.06917)
                (7,1.08973)
                (8,1.25399)
                (9,1.28214)
                (10,1.35425)
                % (11,1.32911)
                % (12,1.45214)
                % (13,1.39665)
                % (14,1.56813)
                % (15,1.68793)
                % (16,1.74085)
                % (17,1.70084)
                % (18,1.73148)
                % (19,1.64343)
                % (20,1.60756)
                % (21,1.72251)
                % (22,1.91925)
                % (23,2.00618)
                % (24,2.09104)
                % (25,1.83422)
            };
            \addlegendentry{KLT}
        \end{axis}
    \end{tikzpicture}
    \caption{Epipolar error statistics on all data from EuRoC. The x-axis represents track length, and the y-axis indicates the median value of the epipolar error corresponding to the track length.}
    \label{fig:Epipolar_Error}
\end{figure}

% \begin{figure}
%     \centering
% \label{Track-Length}
%     \includegraphics[width=1\linewidth]{pictures/Track-Length.png}
%     \caption{Track length statistics for different sequences in EuRoC.}
%     \label{fig:track-length}
% \end{figure}

\begin{figure}[h]
    \centering
    \begin{tikzpicture}
        \begin{axis}[
            ylabel={Track Length},
            legend pos=north east, % 图例位置
            legend columns=3, % 设置图例的列数为3
            grid=both, % 显示网格
            major grid style={dotted,gray}, % 主要网格线样式
            minor grid style={dotted,gray}, % 次要网格线样式
            xmajorgrids=false, % 不显示垂直主要网格线
            xminorgrids=false, % 不显示垂直次要网格线
            width=\linewidth, % 宽度设置为12厘米
            height=0.5\linewidth, % 高度设置为4厘米
            xtick=data, % 使用数据点的值作为横坐标刻度
            xticklabels={MH01, MH02, MH03, MH04, MH05, V101, V102, V103, V201, V202, V203}, % 设置横坐标的标签
            xticklabel style={rotate=90,anchor=east,xshift=0cm}, % 标签旋转为垂直，锚点为东
            ]
            \addplot[color=blue,mark=*] coordinates {
                (1,9.16401)
                (2,8.84333)
                (3,7.94203)
                (4,7.54585)
                (5,7.70342)
                (6,8.44075)
                (7,5.76857)
                (8,4.86837)
                (9,7.20563)
                (10,5.2953)
                (11,4.26726)
            };
            \addlegendentry{Hyb}
            \addplot[color=yellow,mark=*] coordinates {
                (1,4.03109)
                (2,3.77664)
                (3,3.49368)
                (4,3.53465)
                (5,3.49801)
                (6,3.19997)
                (7,2.90238)
                (8,2.83099)
                (9,3.3095)
                (10,2.93061)
                (11,2.80861)
            };
            \addlegendentry{ORB}
            \addplot[color=red,mark=*] coordinates {
                (1,22.1775)
                (2,19.655)
                (3,14.8748)
                (4,16.1651)
                (5,16.8731)
                (6,18.4386)
                (7,8.61279)
                (8,6.32001)
                (9,16.6511)
                (10,8.50012)
                (11,4.93324)
            };
            \addlegendentry{KLT}
        \end{axis}
    \end{tikzpicture}
    \caption{Track length statistics for different sequences in EuRoC. Y-axis indicates the mean track length of each sequence.}
    \label{fig:track-length}
\end{figure}
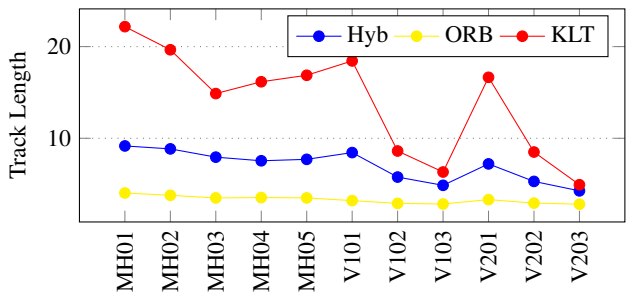

To assess the effectiveness of the hybrid matching method, we conducted a comparative analysis using different matching strategies on the EuRoC dataset. These strategies include ORB's KNN matching, optical flow matching and hybrid matching. 

To evaluate the accuracy of feature matching, we employed the epipolar error metric. Using the true pose values provided by the dataset, we computed the relative pose between the two matched frames. Subsequently, with the relative pose and intrinsic parameters, we derived the fundamental matrix. The epipolar error metric is then calculated using the epipolar geometric relationship as follows:
\begin{equation}
        E_{epi} = \textbf{p}_i^T \textbf{F} \textbf{p}_j,
\end{equation}
where $\textbf{F}$ represents the fundamental matrix between frame $i$ and frame $j$ ,  and $\textbf{p}_i,\textbf{p}_j$ denote the corresponding feature points on the two frames.
We analyzed the relationship between track length and epipolar error across different feature matching methods using the entire EuRoC dataset. Specifically, \textit{ORB} denotes continuous frame KNN matching based on ORB features, \textit{KLT} signifies the optical flow method, and \textit{Hyb} represents our proposed hybrid method. From the results depicted in \cref{fig:Epipolar_Error}, it is observed that as the track length increases, the epipolar error also increases, indicating a positive correlation trend. Descriptor-based matching exhibits the smallest epipolar error, while optical flow-based matching demonstrates the largest epipolar error. Our hybrid matching method lies between them. Compared with optical flow matching, the hybrid method demonstrates a significant improvement in matching accuracy.

We further conducted statistical analysis on the track length of different matching strategies. In \cref{fig:track-length}, we observe that the optical flow-based method consistently yields longer track lengths across all sequences. However, these lengths exhibit significant fluctuations with changes in scene and movement speed. Additionally, as shown in \cref{fig:Epipolar_Error},  the long tracks produced by the KLT method may result in large epipolar errors, thereby not contributing to accuracy improvement. Our hybrid method maintains a moderate track length and exhibits stability across different sequences. Compared to the descriptor-based method, the track length of the hybrid approach shows a significant increase, nearly doubling.

We also compared the time consumption of the hybrid method and the KLT method in \cref{tab:time_statistics}. Despite the hybrid method utilizing both ORB and KLT in feature matching, the overall time consumption does not increase significantly. This is attributed to the shared feature detection module of ORB and KLT within the hybrid method, with the time-consuming increase primarily occurring in the matching process. Furthermore, leveraging the prior knowledge of KLT eliminates the need to extract an excessive number of ORB points (in our case, only 150 points are extracted), which proves adequate for matching within the sliding window. The time cost of \textit{others} includes triangulation, EKF filter, etc. 
\subsection{Trajectory Evaluation}
\begin{table*}[h]
    \caption{ATE (m)  of different algorithms on EuRoC.  Bold font indicates the best result in each column. '-' represents failure to run on this data. All results (except for HybVIO, whose ATE is obtained from \cite{hybvio}) were obtained by ourselves using open-source code and default configurations. None of them incorporate loop closure. }
    \centering
    \begin{tabular}{ccccccccccccc}
    \toprule
         & MH-01 & MH-02 & MH-03 & MH-04 & MH-05 & V1-01 & V1-02 & V1-03 & V2-01 & V2-02 & V2-03 & Avg. \\
         \midrule
         OKVIS      & 0.337 & 0.306 & 0.253 & 0.305 & 0.392 & 0.090 & 0.145 & 0.255 & 0.234 & 0.163 & 0.242 & 0.247 \\
         VINS-Mono  & 0.155 & 0.178 & 0.224 & 0.344 & 0.293 & 0.089 & 0.112 & 0.180 & 0.082 & 0.129 & 0.307 & 0.190 \\
         VINS-Fusion& 0.181 & 0.092 & 0.167 & 0.203 & 0.304 & 0.064 & 0.270 & 0.158 & 0.082 & -     & 0.160 & 0.168 \\
         OpenVINS   & \textbf{0.079} & 0.149 & 0.138 & 0.204 & 0.502 & 0.061 & 0.063 & \textbf{0.063} & 0.101 & \textbf{0.064} & 0.178 & 0.146 \\
         HybVIO     & 0.190 & \textbf{0.066} & \textbf{0.120} & 0.210 & 0.310 & 0.069 & 0.061 & 0.080 & \textbf{0.052} & 0.089 & 0.130 & 0.130 \\
         \midrule
         XR-VIO(KLT)& 0.106 & 0.144 & 0.145 & 0.230 & \textbf{0.212} & \textbf{0.050} & 0.044 & 0.100 & 0.073 & 0.071 & 0.158 & 0.121 \\
         XR-VIO(ORB)& 0.164 & 0.140 & 0.204 & 0.354 & 0.309 & 0.124 & 0.107 & 0.201 & 0.064 & 0.186 & 0.200 & 0.186 \\
         XR-VIO     & 0.082 & 0.114 & 0.164 & \textbf{0.181} & 0.215 & 0.060 & \textbf{0.041} & 0.075 & 0.062 & 0.081 & \textbf{0.100} & \textbf{0.107} \\
         \bottomrule
    \end{tabular}
    \label{tab:ATE_Compare}
\end{table*}
\begin{table*}[h]

    \caption{ATE (mm)  of different algorithms on the ZJU-Sensetime dataset. Bold font indicates the best result in each column. '-' represents failure to run on this data. XR-VIO (KLT) is the ablation version of our XR-VIO, utilizing only KLT for feature matching.}
    \centering
    \begin{tabular}{cccccccccc}
    \toprule
         &  A0&  A1&  A2&  A3&  A4&  A5&  A6&  A7& Avg.\\
         \midrule
         OKVIS\protect\footnotemark[1]          &  71.677 & 87.730 & 68.381 & 22.949 &  146.890 & 77.924 & 63.895 & 47.465 & 73.364 \\
         VINS-Mono\protect\footnotemark[1]      &  63.395 & 80.687 & 74.842 & 19.964 & 18.691 & 42.451 & 26.240 &  18.226 & 43.062 \\
         Vins-Fusion\protect\footnotemark[3]&  77.200 & 134.112 & 40.739 & 23.901 & 28.915 & 57.254 & 26.840 & 25.174 & 51.767\\
         OpenVINS\protect\footnotemark[3]& 74.368 & 97.339 & 41.610 & 22.387 & 16.862 & -   & 22.828 & 15.350 & 41.534\\
         HybVIO\protect\footnotemark[2]         &  \textbf{49.900}&  \textbf{30.000 }&  36.000&  22.200&  19.600&  37.800&  29.300&  17.300& 30.300\\
         SenseSLAM(V1.0)\protect\footnotemark[1]&  58.995 & 55.097 & 36.370 & 17.792 & 15.558 & 34.810 & 20.467 & \textbf{10.777} & 31.233\\
\hline
         XR-VIO(KLT)\protect\footnotemark[3]    &  \phantom{1}60.385\phantom{1} &  \phantom{1}71.926\phantom{1} &  \phantom{1}31.134\phantom{1} &  \phantom{1}16.657\phantom{1} &  \phantom{1}25.913\phantom{1} &  \phantom{1}34.288\phantom{1} &  \phantom{1}20.410\phantom{1} &  \phantom{1}13.125\phantom{1} & \phantom{1}34.230\phantom{1} \\
         XR-VIO(ORB)\protect\footnotemark[3]& 71.694 & 66.478 & 45.526 & 17.621 & 31.178 & 48.582 & 20.355 & 22.422 & 40.482\\
        XR-VIO\protect\footnotemark[3]          &  56.604 &  46.219 & \textbf{30.422} & \textbf{15.291}& \textbf{15.078}&  \textbf{30.283}& \textbf{17.082}&  12.598 & \textbf{27.947}\\
        \bottomrule
    \end{tabular}
    
    \label{tab:ATE_ZJU}
       \footnotemark[1]{ATE is obtained from \cite{jinyu2019survey}  }
       \footnotemark[2]{ATE is obtained from \cite{hybvio} }
       \footnotemark[3]{ATE is obtained by ourselves.}
\end{table*}

In \cref{tab:ATE_Compare,tab:ATE_ZJU}, we respectively compared our approach with the current state-of-the-art \cite{leutenegger-ijrr-2015-OKVIS, qin-tro-2018_VINS-Mono, qin2019a_VINS_Fusion_Local, geneva2020openvins,hybvio} on the EuRoC benchmark and the ZJU-Sensetime benchmark.
% \textit{All of these methods and our own run on same environment. Trajectories are without initialization part.}
% Our approach yields state-of-the-art performance on each dataset.
We also compared our own method with a baseline version employing solely KLT and ORB methods for feature matching, illustrating  the efficiency of our hybrid feature matching. The results demonstrate a noticeable improvement in ATE on both the EuRoC and ZJU-Sensetime datasets.

The ZJU-Sensetime dataset comprises data from consumer-grade mobile phones, captured with rolling cameras resulting in low-quality images often affected by motion blur. As a result, the accuracy of most algorithms is notably reduced when applied to this dataset. Our hybrid matching strategy demonstrates its advantage, particularly under the conditions present in the mobile phone dataset. Optical flow-based methods are particularly sensitive to variations in image quality, with rapid motion or changes in brightness causing significant errors in feature matching. However, our hybrid matching strategy effectively mitigates these issues, thereby improving the accuracy of VIO.
Additionally, more experimental results for accuracy comparison, such as evaluation on ADVIO \cite{cortes2018advio} dataset, can be found in our supplementary material.

In addition to the methods outlined in \cref{tab:ATE_Compare,tab:ATE_ZJU} , we refrain from comparing several other VIO/VI-SLAM systems, such as VI-ORB \cite{murartal-ral-2017-VI-ORB}, ORB-SLAM3 \cite{campos2021orb-slam3}, VI-DSO \cite{von2018direct-VI-DSO}, DM-VIO \cite{stumberg22DM-VIO}, despite their demonstrated high accuracy. This decision stems from their inherent latency issues, rendering them unsuitable for XR applications Latency manifests in two main aspects. Firstly, there is the computational burden. While algorithms in the ORB-SLAM series offer good accuracy, they rely on a significant number of complex calculations, making deployment on mobile platforms challenging. Secondly, there are delays inherent in the algorithm mechanisms. For instance, the ORB-SLAM\cite{murartal-ral-2017-VI-ORB, campos2021orb-slam3} series require 15 seconds of data to complete initialization, and DSO \cite{von2018direct-VI-DSO, stumberg22DM-VIO} series, with delayed marginalization, also need 2\textasciitilde10 seconds to obtain real scale. These latency issues are unacceptable for XR users.
\subsection{Ablation Study}
We conducted an ablation study on each initialization module under the 4KF configuration, as presented in \cref{tab:Ablation}. This analysis aimed to assess the individual impact of each module on initialization performance. The left column lists different algorithm strategies. \textit{Full} denotes our complete strategy for initialization, where all indicators demonstrate the best performance. Subsequently, we examined various ablation strategies. The ablation of \textit{2-point} demonstrates an improvement in the success rate of initialization. This method, when combined with known rotation, simplifies the solution for the initial relative pose. In contrast, the \textit{5-point} method requires solving complex polynomial problems and selecting the correct solution from a candidate set. \textit{VG-BA} represents an enhancement of traditional Visual-BA. By incorporating the orientation constraint of the gyroscope, the solution of the entire SfM problem becomes less prone to divergence, especially in cases of small motion and parallax. Moreover, VG-BA improves the accuracy of orientation estimation by optimizing gyroscope bias. Following VG-BA and initial visual-inertial alignment, the overall VI-BA plays a crucial role. The last two rows of \cref{tab:Ablation} demonstrate the impact of missing complete or partial VI-BA on initialization accuracy. Without VI-BA, scale error significantly increases, highlighting the importance of the weighting of visual terms.
\begin{table}[h]
    \centering

    \caption{Ablation study for VI-Initialization. Bold font indicates the best result, and Italic font highlights significant impact resulting from ablation.}
    \begin{tabular}{ccccc}
    \toprule
          &  Scale(\%)$\downarrow$&  ATE(m)$\downarrow$&  Gravity(°)$\downarrow$&  Succ(\%)$\uparrow$\\
          \midrule
           Full&  \textbf{26.88}&  \textbf{0.026}&  \textbf{2.26}&  \textbf{83.98}\\
           w/o 2-point&  27.08&  0.027&  2.34&  \textit{\textbf{80.43}}\\
           w/o VG-BA&  26.92&  0.027&  \textit{\textbf{2.45}} &  82.62\\
           w/o VI-BA&  \textit{\textbf{30.68}}&  0.028&  2.37&  83.98\\
          w/o weight&  \textit{\textbf{28.27}}&  0.028&  2.30&  83.84\\
    \bottomrule
    \end{tabular}
    \label{tab:Ablation}
\end{table}

\subsection{Mobile AR}
\begin{figure}
    \centering
    \includegraphics[width=\columnwidth]{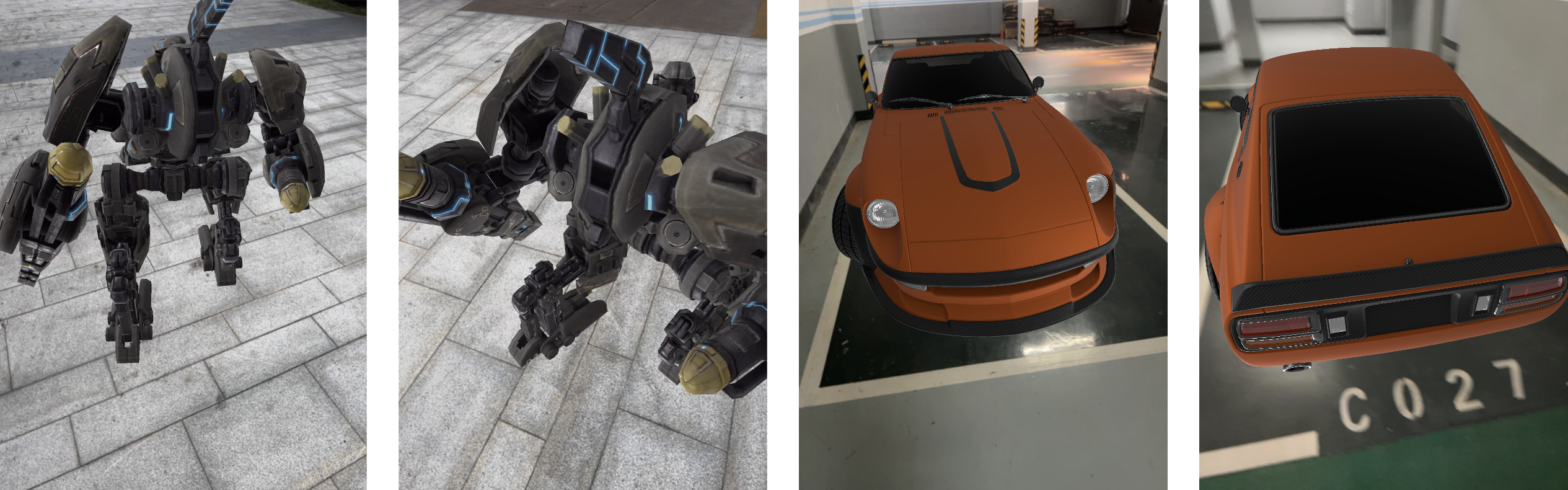}
    \caption{AR effects on mobile platforms}
    \label{fig:fancy AR demo}
\end{figure}

%We deployed our VIO system to mobile platforms (Android \& iOS) and developed a simple AR demo to demonstrate its accuracy and robustness. We utilized 30 Hz image data with a resolution of $640\times480$ and IMU data, including angular velocity and acceleration captured at 100/200Hz by the iPhone 13 Pro and Huawei Mate30 Pro. Our system operates in real-time on mobile devices, seamlessly integrating virtual objects into real-world scenes. \cref{fig:fancy AR demo} shows two AR examples. We assessed the accuracy of the algorithm by examining loop errors in the 3D trajectory during long-distance motion tracking in expansive scenes. Over tracking distances of several hundred meters, no visually significant loop errors or height discrepancies were observed. The experimental results demonstrate the superiority of our system in both initialization and normal tracking tasks, highlighting the effectiveness of our algorithm in XR applications. Further details of our mobile AR implementation can be found in our supplementary video material.
We deployed our VIO system on mobile platforms (Android \& iOS) and created a simple AR demo to showcase its accuracy and robustness. Using 30 Hz image data at a resolution of 640×480 and IMU data, including angular velocity and acceleration captured at 100/200Hz by the iPhone 13 Pro and Huawei Mate30 Pro, our system operates in real-time on mobile devices. It seamlessly integrates virtual objects into real-world scenes. \cref{fig:fancy AR demo} displays two AR examples. We evaluated the algorithm's accuracy by examining loop errors in the 3D trajectory during long-distance motion tracking in large scenes. Over distances of several hundred meters, no significant loop errors or height discrepancies were visually observed. The experimental results highlight the system's superior performance in both initialization and normal tracking tasks, emphasizing the algorithm's effectiveness in XR applications. Additional details of our mobile AR implementation are available in our supplementary video material.
% \begin{figure}
%     \centering
%     \includegraphics[width=1\columnwidth]{pictures/Mobile-Trajectory.png}
%     \caption{Trajectory of VIO on a mobile phone: \textcolor{red}{need to add pic @nan, online AR + Trajectory}.}
%     \label{fig:fancy AR demo}
% \end{figure}

\section{Conclusion}
In this paper, we propose a novel and robust VIO system capable of supporting rapid visual-inertial initialization and precise 6DoF tracking. For initialization, we propose VG-SfM, which significantly enhances the robustness of camera pose and gyroscope bias estimation. In terms of feature matching, we propose a hybrid scheme that tightly integrates optical flow and descriptor-based matching, enabling feature tracking to maintain high accuracy even with long track length.  Both approaches contribute significantly to the performance enhancement of our algorithm. 

Our experimental results on the EuRoC, ZJU-Sensetime and ADVIO datasets demonstrate clear superiority over existing methods, validating the effectiveness of our system. Moreover, our algorithm exhibits good computational efficiency, enabling real-time execution on mobile devices. An AR demo on mobile devices further showcases the algorithm's robustness in challenging scenarios and highlights its potential for mobile AR applications.

% TODO
Despite these achievements, our system still faces limitations in extremely challenging environments such as pure-rotation scenes, textureless environments, and dynamic environments. Furthermore, we acknowledge that deep learning-based methods hold promise for further improving VIO performance. For instance, incorporating learning-based depth estimation and inertial navigation systems could provide additional geometric and motion measurements suitable for VIO or SLAM. We plan to explore these avenues in future work to refine and enhance our algorithm.

% The authors wish to thank A, B, and C. This work was supported in part by
% a grant from XYZ.}

%\bibliographystyle{abbrv}
%\bibliographystyle{abbrv_conference/abbrv-doi}
\bibliographystyle{abbrv_journal/abbrv-doi}

\bibliography{main}

\begin{thebibliography}{10}

\bibitem{bang2017camera}
J.~Bang, D.~Lee, Y.~Kim, and H.~Lee.
\newblock Camera pose estimation using optical flow and orb descriptor in
  slam-based mobile ar game.
\newblock In {\em 2017 International Conference on Platform Technology and
  Service (PlatCon)}, pp. 1--4. IEEE, 2017.

\bibitem{bao2022robust}
H.~Bao, W.~Xie, Q.~Qian, D.~Chen, S.~Zhai, N.~Wang, and G.~Zhang.
\newblock Robust tightly-coupled visual-inertial odometry with pre-built maps
  in high latency situations.
\newblock {\em IEEE Transactions on Visualization and Computer Graphics},
  28(5):2212--2222, 2022.

\bibitem{bloesch-ijrr-2017}
M.~Bloesch, M.~Burri, S.~Omari, M.~Hutter, and R.~Siegwart.
\newblock Iterated extended kalman filter based visual-inertial odometry using
  direct photometric feedback.
\newblock {\em The International Journal of Robotics Research},
  36(10):1053--1072, 2017.

\bibitem{opencv_library}
G.~Bradski.
\newblock {The OpenCV Library}.
\newblock {\em Dr. Dobb's Journal of Software Tools}, 2000.

\bibitem{Burri25012016-EuRoC}
M.~Burri, J.~Nikolic, P.~Gohl, T.~Schneider, J.~Rehder, S.~Omari, M.~W.
  Achtelik, and R.~Siegwart.
\newblock The {EuRoC} micro aerial vehicle datasets.
\newblock {\em The International Journal of Robotics Research}, 2016.

\bibitem{Cai_2021-pose-only}
Q.~Cai, L.~Zhang, Y.~Wu, W.~Yu, and D.~Hu.
\newblock A pose-only solution to visual reconstruction and navigation.
\newblock {\em IEEE Transactions on Pattern Analysis and Machine Intelligence},
  45(1):73--86, 2021.

\bibitem{calonder2010brief}
M.~Calonder, V.~Lepetit, C.~Strecha, and P.~Fua.
\newblock Brief: Binary robust independent elementary features.
\newblock In {\em Computer Vision--ECCV 2010: 11th European Conference on
  Computer Vision, Heraklion, Crete, Greece, September 5-11, 2010, Proceedings,
  Part IV 11}, pp. 778--792. Springer, 2010.

\bibitem{campos2021orb-slam3}
C.~Campos, R.~Elvira, J.~J.~G. Rodr{\'\i}guez, J.~M. Montiel, and J.~D.
  Tard{\'o}s.
\newblock {ORB-SLAM3}: An accurate open-source library for visual,
  visual--inertial, and multimap {SLAM}.
\newblock {\em IEEE Transactions on Robotics}, 37(6):1874--1890, 2021.

\bibitem{campos2019fast}
C.~Campos, J.~M. Montiel, and J.~D. Tard{\'o}s.
\newblock Fast and robust initialization for visual-inertial slam.
\newblock In {\em 2019 International Conference on Robotics and Automation
  (ICRA)}, pp. 1288--1294. IEEE, 2019.

\bibitem{campos2020inertial}
C.~Campos, J.~M. Montiel, and J.~D. Tard{\'o}s.
\newblock Inertial-only optimization for visual-inertial initialization.
\newblock In {\em 2020 IEEE International Conference on Robotics and Automation
  (ICRA)}, pp. 51--57. IEEE, 2020.

\bibitem{chen2021rnin}
D.~Chen, N.~Wang, R.~Xu, W.~Xie, H.~Bao, and G.~Zhang.
\newblock {RNIN-VIO}: Robust neural inertial navigation aided visual-inertial
  odometry in challenging scenes.
\newblock In {\em 2021 IEEE International Symposium on Mixed and Augmented
  Reality (ISMAR)}, pp. 275--283. IEEE, 2021.

\bibitem{cortes2018advio}
S.~Cort{\'e}s, A.~Solin, E.~Rahtu, and J.~Kannala.
\newblock {ADVIO}: An authentic dataset for visual-inertial odometry.
\newblock In {\em Proceedings of the European Conference on Computer Vision
  (ECCV)}, pp. 419--434, 2018.

\bibitem{Dong-Si_initialization}
T.-C. Dong-Si and A.~I. Mourikis.
\newblock Estimator initialization in vision-aided inertial navigation with
  unknown camera-imu calibration.
\newblock In {\em 2012 IEEE/RSJ International Conference on Intelligent Robots
  and Systems}, pp. 1064--1071, 2012. doi: {{%
10\hspace{.1pt}\discretionary{.}{%
}{.}\hspace{.4pt}1109\discretionary{/}{%
}{/}IROS\hspace{.1pt}\discretionary{.}{%
}{.}\hspace{.4pt}2012\hspace{.1pt}\discretionary{.}{%
}{.}\hspace{.4pt}6386235}}


\bibitem{engel2018direct_DSO}
J.~Engel, V.~Koltun, and D.~Cremers.
\newblock Direct sparse odometry.
\newblock {\em IEEE transactions on pattern analysis and machine intelligence},
  40(3):611--625, 2018.

\bibitem{forster2017manifold-preintergration}
C.~Forster, L.~Carlone, F.~Dellaert, and D.~Scaramuzza.
\newblock On-manifold preintegration for real-time visual--inertial odometry.
\newblock {\em IEEE Transactions on Robotics}, 33(1):1--21, 2017.

\bibitem{geneva2020openvins}
P.~Geneva, K.~Eckenhoff, W.~Lee, Y.~Yang, and G.~Huang.
\newblock {OpenVINS}: A research platform for visual-inertial estimation.
\newblock In {\em 2020 IEEE International Conference on Robotics and Automation
  (ICRA)}, pp. 4666--4672. IEEE, 2020.

\bibitem{genevaopenvins}
P.~Geneva and G.~Huang.
\newblock Openvins state initialization: Details and derivations.
\newblock Technical report, Tech. Rep. RPNG-2022-INIT, University of Delaware,
  2022.

\bibitem{grupp2017evo}
M.~Grupp.
\newblock evo: Python package for the evaluation of odometry and slam.
\newblock \url{https://github.com/MichaelGrupp/evo}, 2017.

\bibitem{Rotation-Translation-Decoupled}
Y.~He, B.~Xu, and H.~Li.
\newblock A rotation-translation-decoupled solution for robust and efficient
  visual-inertial initialization.
\newblock In {\em Proceedings of the IEEE/CVF Conference on Computer Vision and
  Pattern Recognition}, pp. 739--748, 2023.

\bibitem{huai2022robocentric}
Z.~Huai and G.~Huang.
\newblock Robocentric visual-inertial odometry.
\newblock {\em The International Journal of Robotics Research}, 41(7):667--689,
  2022.

\bibitem{jinyu2019survey}
L.~Jinyu, Y.~Bangbang, C.~Danpeng, W.~Nan, Z.~Guofeng, and B.~Hujun.
\newblock Survey and evaluation of monocular visual-inertial {SLAM} algorithms
  for augmented reality.
\newblock {\em Virtual Reality \& Intelligent Hardware}, 1(4):386--410, 2019.

\bibitem{Kneip_Chli_Siegwart_2010}
L.~Kneip, M.~Chli, and R.~Siegwart.
\newblock Robust real-time visual odometry with a single camera and an imu.
\newblock In {\em Procedings of the British Machine Vision Conference 2011},
  Dec 2010. doi: {{%
10\hspace{.1pt}\discretionary{.}{%
}{.}\hspace{.4pt}5244\discretionary{/}{%
}{/}c\hspace{.1pt}\discretionary{.}{%
}{.}\hspace{.4pt}25\hspace{.1pt}\discretionary{.}{%
}{.}\hspace{.4pt}16}}


\bibitem{Kneip_Lynen_2013-rotation}
L.~Kneip and S.~Lynen.
\newblock Direct optimization of frame-to-frame rotation.
\newblock In {\em 2013 IEEE International Conference on Computer Vision}, Nov
  2013. doi: {{%
10\hspace{.1pt}\discretionary{.}{%
}{.}\hspace{.4pt}1109\discretionary{/}{%
}{/}iccv\hspace{.1pt}\discretionary{.}{%
}{.}\hspace{.4pt}2013\hspace{.1pt}\discretionary{.}{%
}{.}\hspace{.4pt}292}}


\bibitem{BRISK}
S.~Leutenegger, M.~Chli, and R.~Y. Siegwart.
\newblock Brisk: Binary robust invariant scalable keypoints.
\newblock In {\em 2011 International Conference on Computer Vision}, pp.
  2548--2555, 2011. doi: {{%
10\hspace{.1pt}\discretionary{.}{%
}{.}\hspace{.4pt}1109\discretionary{/}{%
}{/}ICCV\hspace{.1pt}\discretionary{.}{%
}{.}\hspace{.4pt}2011\hspace{.1pt}\discretionary{.}{%
}{.}\hspace{.4pt}6126542}}


\bibitem{leutenegger-ijrr-2015-OKVIS}
S.~Leutenegger, S.~Lynen, M.~Bosse, R.~Siegwart, and P.~Furgale.
\newblock Keyframe-based visual–inertial odometry using nonlinear
  optimization.
\newblock {\em The International Journal of Robotics Research}, 34(3):314--334,
  2015.

\bibitem{Lucas_Kanade_1981_KLT}
B.~Lucas and T.~Kanade.
\newblock An iterative image registration technique with an application to
  stereo vision.
\newblock Aug 1981.

\bibitem{martinelli2014closed}
A.~Martinelli.
\newblock Closed-form solution of visual-inertial structure from motion.
\newblock {\em International journal of computer vision}, 106(2):138--152,
  2014.

\bibitem{maybeck1982stochastic-square-root}
P.~S. Maybeck.
\newblock {\em Stochastic models, estimation, and control}.
\newblock Academic press, 1982.

\bibitem{merrill2023fast}
N.~Merrill, P.~Geneva, and S.~K. C. C.~G. Huang.
\newblock Fast monocular visual-inertial initialization leveraging learned
  single-view depth.
\newblock In {\em Robotics: Science and Systems (RSS)}, vol.~2, 2023.

\bibitem{mourikis-icra-2007}
A.~I. Mourikis and S.~I. Roumeliotis.
\newblock A multi-state constraint kalman filter for vision-aided inertial
  navigation.
\newblock In {\em Proceedings 2007 IEEE International Conference on Robotics
  and Automation}, pp. 3565--3572, 4 2007.

\bibitem{murartal-tro-2015-ORB-SLAM}
R.~Mur-Artal, J.~M.~M. Montiel, and J.~D. Tard{\'o}s.
\newblock {ORB-SLAM}: a versatile and accurate monocular {SLAM} system.
\newblock {\em IEEE Transactions on Robotics}, 31(5):1147--1163, 2015.

\bibitem{murartal-ral-2017-VI-ORB}
R.~Mur-Artal and J.~D. Tard{\'o}s.
\newblock Visual-inertial monocular {SLAM} with map reuse.
\newblock {\em IEEE Robotics and Automation Letters}, 2(2):796--803, 2017.

\bibitem{fivepoint2004nister}
D.~Nister.
\newblock An efficient solution to the five-point relative pose problem.
\newblock {\em IEEE Transactions on Pattern Analysis and Machine Intelligence},
  26(6):756--770, June 2004.

\bibitem{qin2019b_VINS_Fusion_Global}
T.~Qin, S.~Cao, J.~Pan, and S.~Shen.
\newblock A general optimization-based framework for global pose estimation
  with multiple sensors, 2019.

\bibitem{qin-tro-2018_VINS-Mono}
T.~Qin, P.~Li, and S.~Shen.
\newblock {VINS-Mono}: A robust and versatile monocular visual-inertial state
  estimator.
\newblock {\em IEEE Transactions on Robotics}, 34(4):1004--1020, 2018.

\bibitem{qin2019a_VINS_Fusion_Local}
T.~Qin, J.~Pan, S.~Cao, and S.~Shen.
\newblock A general optimization-based framework for local odometry estimation
  with multiple sensors, 2019.

\bibitem{Qin_Shen_2017}
T.~Qin and S.~Shen.
\newblock Robust initialization of monocular visual-inertial estimation on
  aerial robots.
\newblock In {\em 2017 IEEE/RSJ International Conference on Intelligent Robots
  and Systems (IROS)}, Sep 2017. doi: {{%
10\hspace{.1pt}\discretionary{.}{%
}{.}\hspace{.4pt}1109\discretionary{/}{%
}{/}iros\hspace{.1pt}\discretionary{.}{%
}{.}\hspace{.4pt}2017\hspace{.1pt}\discretionary{.}{%
}{.}\hspace{.4pt}8206284}}


\bibitem{ranftl2020towards}
R.~Ranftl, K.~Lasinger, D.~Hafner, K.~Schindler, and V.~Koltun.
\newblock Towards robust monocular depth estimation: Mixing datasets for
  zero-shot cross-dataset transfer.
\newblock {\em IEEE transactions on pattern analysis and machine intelligence},
  44(3):1623--1637, 2020.

\bibitem{rublee2011orb}
E.~Rublee, V.~Rabaud, K.~Konolige, and G.~Bradski.
\newblock Orb: An efficient alternative to sift or surf.
\newblock In {\em 2011 International conference on computer vision}, pp.
  2564--2571. Ieee, 2011.

\bibitem{sarlin2020superglue}
P.-E. Sarlin, D.~DeTone, T.~Malisiewicz, and A.~Rabinovich.
\newblock Superglue: Learning feature matching with graph neural networks.
\newblock In {\em Proceedings of the IEEE/CVF conference on computer vision and
  pattern recognition}, pp. 4938--4947, 2020.

\bibitem{Schinazi2022-CDF}
R.~B. Schinazi.
\newblock {\em The Cumulative Distribution Function}, pp. 159--168.
\newblock Springer International Publishing, Cham, 2022. doi: {{%
10\hspace{.1pt}\discretionary{.}{%
}{.}\hspace{.4pt}1007\discretionary{/}{%
}{/}978\discretionary{%
}{-}{-}3\discretionary{%
}{-}{-}030\discretionary{%
}{-}{-}93635\discretionary{%
}{-}{-}8\_15}}


\bibitem{hybvio}
O.~Seiskari, P.~Rantalankila, J.~Kannala, J.~Ylilammi, E.~Rahtu, and A.~Solin.
\newblock Hybvio: Pushing the limits of real-time visual-inertial odometry.
\newblock In {\em Proceedings of the IEEE/CVF Winter Conference on Applications
  of Computer Vision}, pp. 701--710, 2022.

\bibitem{gftt1994shi}
J.~Shi and Tomasi.
\newblock Good features to track.
\newblock In {\em 1994 Proceedings of IEEE Conference on Computer Vision and
  Pattern Recognition}, pp. 593--600, June 1994.

\bibitem{sun2021loftr}
J.~Sun, Z.~Shen, Y.~Wang, H.~Bao, and X.~Zhou.
\newblock Loftr: Detector-free local feature matching with transformers.
\newblock In {\em Proceedings of the IEEE/CVF conference on computer vision and
  pattern recognition}, pp. 8922--8931, 2021.

\bibitem{triggs1999bundle}
B.~Triggs, P.~F. McLauchlan, R.~I. Hartley, and A.~W. Fitzgibbon.
\newblock Bundle adjustment—a modern synthesis.
\newblock In {\em International workshop on vision algorithms}, pp. 298--372.
  Springer, 1999.

\bibitem{stumberg22DM-VIO}
L.~von Stumberg and D.~Cremers.
\newblock {DM-VIO}: Delayed marginalization visual-inertial odometry.
\newblock {\em {IEEE} Robotics and Automation Letters ({RA-L})},
  7(2):1408--1415, 2022.

\bibitem{von2018direct-VI-DSO}
L.~Von~Stumberg, V.~Usenko, and D.~Cremers.
\newblock Direct sparse visual-inertial odometry using dynamic marginalization.
\newblock In {\em 2018 IEEE International Conference on Robotics and Automation
  (ICRA)}, pp. 2510--2517. IEEE, 2018.

\bibitem{wu2015square}
K.~Wu, A.~M. Ahmed, G.~A. Georgiou, and S.~I. Roumeliotis.
\newblock A square root inverse filter for efficient vision-aided inertial
  navigation on mobile devices.
\newblock In {\em Robotics: Science and Systems}, vol.~2, p.~2. Rome, Italy,
  2015.

\bibitem{Zhang18iros-Quantitative-Trajectory-Evaluation}
Z.~Zhang and D.~Scaramuzza.
\newblock A tutorial on quantitative trajectory evaluation for
  visual(-inertial) odometry.
\newblock In {\em IEEE/RSJ Int. Conf. Intell. Robot. Syst. (IROS)}, 2018.

\bibitem{zhong2023improved}
L.~Zhong, L.~Meng, W.~Hou, and L.~Huang.
\newblock An improved visual odometer based on lucas-kanade optical flow and
  orb feature.
\newblock {\em IEEE Access}, 11:47179--47186, 2023.

\bibitem{Mono-Depth-VI-Init-2022}
Y.~Zhou, A.~Kar, E.~Turner, A.~Kowdle, C.~Guo, R.~DuToit, and K.~Tsotsos.
\newblock Learned monocular depth priors in visual-inertial initialization.
\newblock Apr 2022.

\bibitem{zong2017improved}
L.~Zong, H.~Wang, B.~Wang, Q.~Fu, and X.~Sun.
\newblock An improved method of real-time camera pose estimation based on
  descriptor tracking.
\newblock In {\em 2017 IEEE 20th International Conference on Intelligent
  Transportation Systems (ITSC)}, pp. 903--908. IEEE, 2017.

\bibitem{zuniga2021analytical}
D.~Zu{\~n}iga-No{\"e}l, F.-A. Moreno, and J.~Gonzalez-Jimenez.
\newblock An analytical solution to the imu initialization problem for
  visual-inertial systems.
\newblock {\em IEEE Robotics and Automation Letters}, 6(3):6116--6122, 2021.

\end{thebibliography}
\end{document}

\firstsection{Introduction}
\maketitle

This is the supplementary document that our main paper refers to. It contains some additional implement details and additional experiment results.

\section{Additional Experiments}

In Sec. 6, we conducted comprehensive experiments on our initialization method and feature tracking approach, comparing them with numerous state-of-the-art (SOTA) algorithms to demonstrate the effectiveness of our proposed method. To further emphasize the advantages of our algorithm, we provide additional experimental results and comparisons in this supplementary material. 

\subsection{Initialization Evaluation}
In Sect. 6.3, we conducted a comprehensive evaluation of our initialization algorithm compared to several SOTA algorithms, including Closed-form \cite{martinelli2014closed}, Vins-Mono \cite{qin-tro-2018_VINS-Mono}, Inertial-only \cite{campos2020inertial}, and DRT \cite{Rotation-Translation-Decoupled}. As presented in Tab. 2 in the main text, we compared the accuracy metrics of our initialization method under two configurations: 4-keyframe (4KF) and 5-keyframe (5KF). Our results demonstrated a high initialization success rate with very few initial frames, outperforming SOTA algorithms in terms of accuracy metrics. However, existing methods typically initialize using a 10-keyframe (10KF) configuration. To provide a more comprehensive and fair comparison, we extended our evaluations to include the 10KF configuration. Under this setting, each sequence was segmented into 1137 fragments at intervals of 1.2 seconds, with keyframes uniformly selected at intervals of 0.1 seconds for all methods.

\begin{figure}[!h]
    \centering
    \includegraphics[width=1\linewidth,height=0.032\linewidth]{pictures/labels.png}
    10KF
    \includegraphics[width=1\linewidth,height=0.3\linewidth]{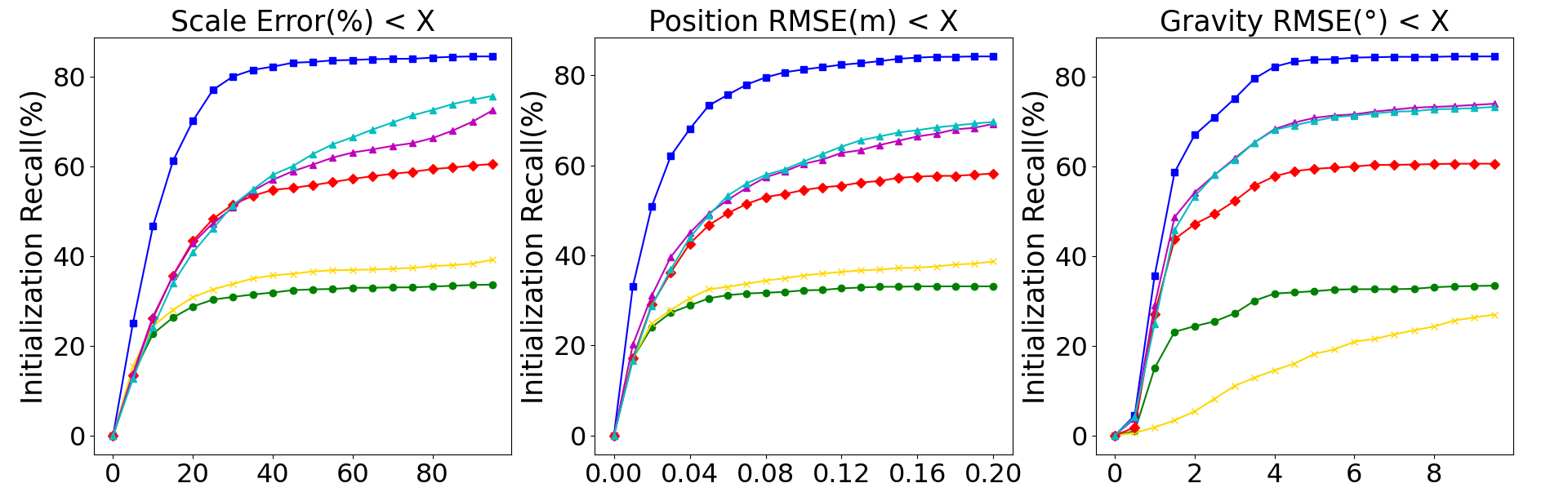}
    \caption{Cumulative distribution of initialization with 10KF. Scale error, ATE and gravity RMSE are shown in 3 columns. }
    \label{fig:Init_Error_10KF_CDF_sup}
\end{figure}

\Cref{tab:Init_Compare2_sup} and  \cref{fig:Init_Error_10KF_CDF_sup} present a detailed overview of the accuracy metrics and corresponding cumulative distribution functions (CDF) of various initialization methods. Notably, our algorithm maintains superiority over existing approaches under the widely adopted 10KF configuration, demonstrating superior performance across all metrics compared to current initialization methods.

Furthermore, it's important to highlight that the DRT method, including DRT-l and DRT-r, does not employ a joint optimization algorithm like VI-BA. To ensure a fair comparison, we conducted additional experiments by disabling VI-BA in the XR-VIO initialization pipeline and compared the results of VA-Align (referred to as XR-VIO w/o VI-BA) with those of the DRT algorithm.

\begin{figure}[!h]
    \centering
    \includegraphics[width=1\linewidth,height=0.05\linewidth]{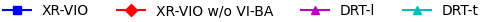}
    4KF
    \centering
    \includegraphics[width=1\linewidth,height=0.3\linewidth]{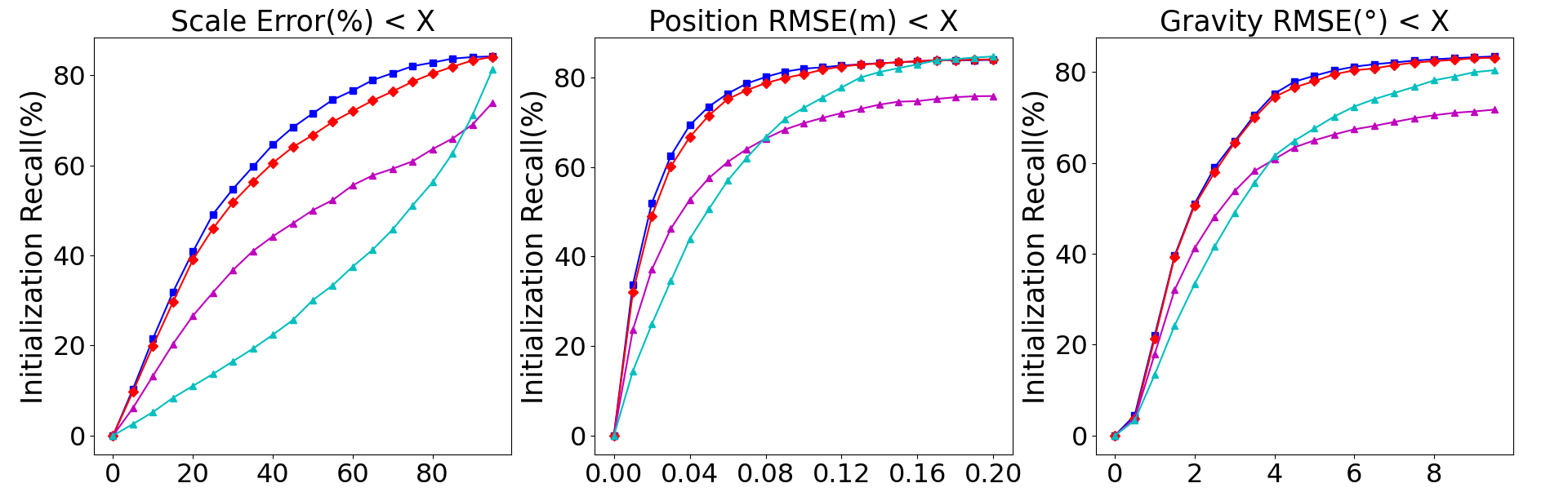}
    5KF
    \includegraphics[width=1\linewidth,height=0.3\linewidth]{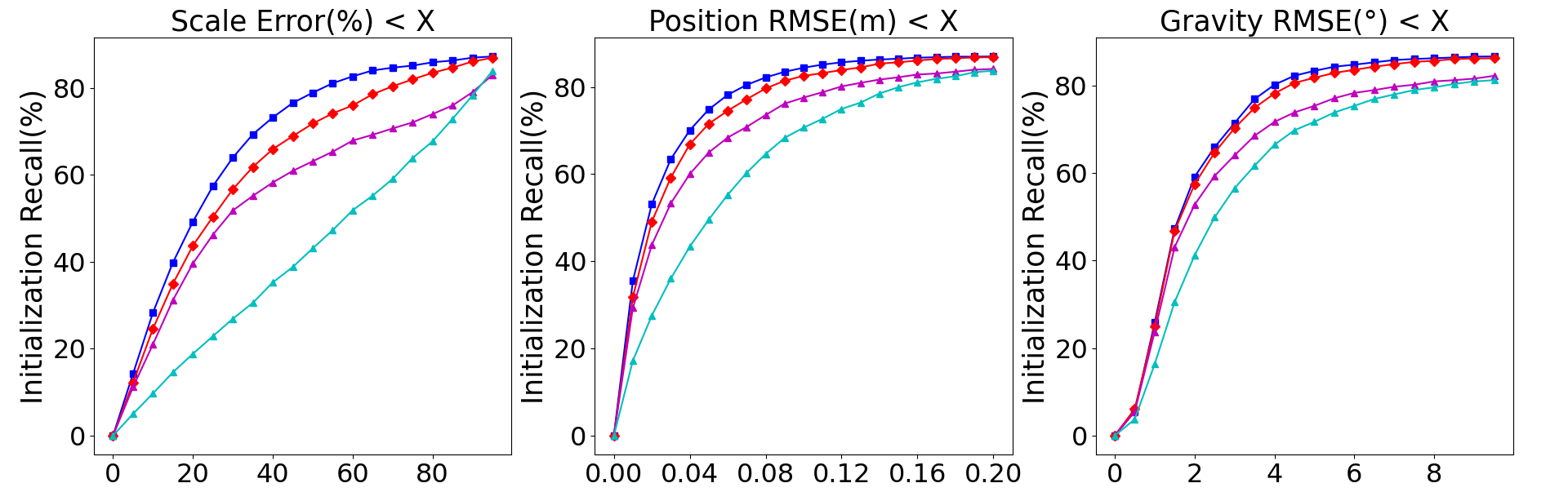}
    10KF
    \includegraphics[width=1\linewidth,height=0.3\linewidth]{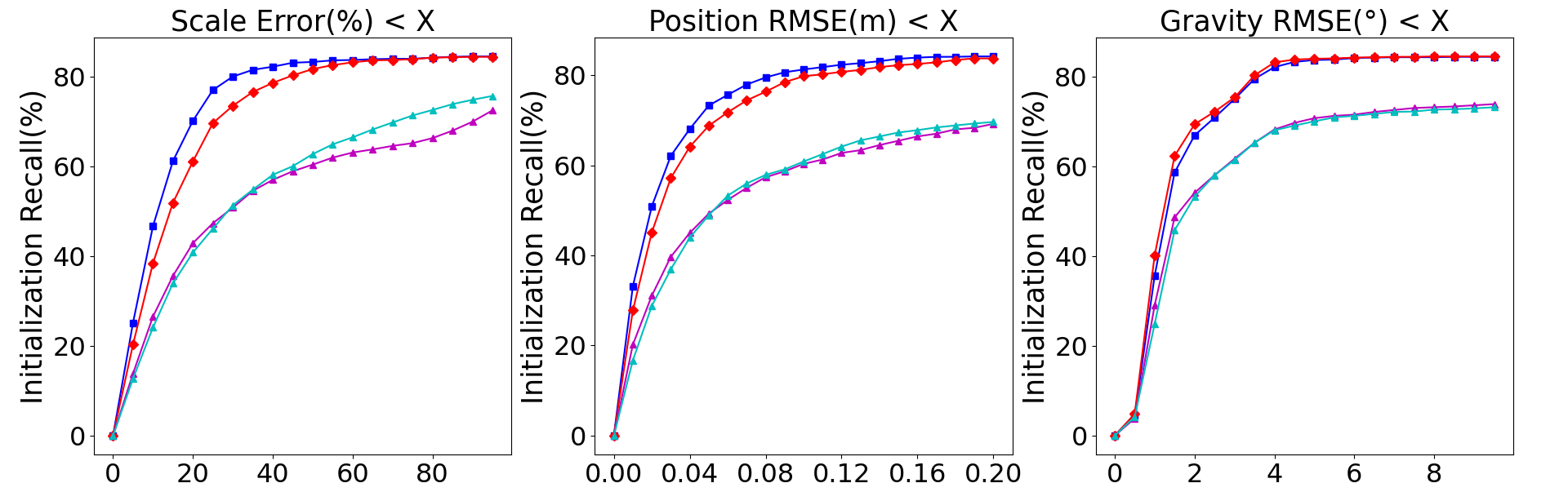}
    \caption{Cumulative distribution of initialization with different keyframes: 4KF, 5KF, and 10KF. Scale error, ATE and gravity RMSE are shown in 3 columns. }
    \label{fig:Init_Error_CDF_sup}
\end{figure}

As illustrated in \cref{fig:Init_Error_CDF_sup} and \cref{tab:Init_Compare_sup}, the accuracy metrics of XR-VIO w/o VI-BA are slightly lower than those of XR-VIO, but significantly higher than those of the DRT algorithm. This result highlights the innovativeness and practical value of our VG-SfM approach, confirming its effectiveness even in scenarios where VI-BA is disabled.

\begin{table}[htbp]
    \centering
    \begin{tabular}{ccccc}
    \toprule
         Algorithms &  Scale(\%)$\downarrow$&  ATE(m)$\downarrow$&  Gravity(°)$\downarrow$&Success(\%)$\uparrow$\\
         \midrule
         Closed-form \cite{martinelli2014closed} & 12.55 & 0.031 & 1.70 & 32.36  \\
         Inertial-only \cite{campos2020inertial} & 20.12 &  0.062& 9.77 & 41.86 \\
         Vins-Mono \cite{qin-tro-2018_VINS-Mono} & 19.78 & 0.053 & 1.72 & 58.51 \\
         DRT-l \cite{Rotation-Translation-Decoupled} & 30.42 & 0.070 & 2.22 & 76.18 \\
         DRT-t \cite{Rotation-Translation-Decoupled} & 27.08 & 0.067 & 2.36 & 75.99 \\
         XR-VIO&  \textbf{11.99}&  \textbf{0.027}&  \textbf{1.48}&\textbf{83.97} \\
         \bottomrule
    \end{tabular}
    \caption{10KF Initialization Evaluation on EuRoC. Bold font indicates the best result. We compare these methods under the configuration: 10KF, including scale, ATE, gravity, and success rate.}
    \label{tab:Init_Compare2_sup}
\end{table}

\begin{table*}[t]
    \centering
    \begin{tabular}{c|cccc|cccc}
    \toprule
 & \multicolumn{4}{c}{4KF} &\multicolumn{4}{c}{5KF}\\
          \midrule
         &  Scale(\%)$\downarrow$&  ATE(m)$\downarrow$&  Gravity(°)$\downarrow$&Success(\%)$\uparrow$& Scale(\%)$\downarrow$& ATE(m)$\downarrow$& Gravity(°)$\downarrow$& Success(\%)$\uparrow$\\
         % \hline
         DRT-l&  41.79&  0.041&  3.50&76.94 & 34.57& 0.039& 2.64&85.79\\
         DRT-t&  61.54&  0.059&  3.97&\textbf{86.56} & 50.46& 0.059& 3.37& \underline{86.24}\\
         XR-VIO w/o VI-BA &  \underline{30.68}&  \underline{0.028}&  \underline{2.37}&\underline{83.98} & \underline{28.34}& \underline{0.032}& \underline{2.14}&\textbf{87.15}\\
         XR-VIO&  \textbf{26.88}&  \textbf{0.026}&  \textbf{2.26}&\underline{83.98} & \textbf{22.71}& \textbf{0.027}& \textbf{1.99}&\textbf{87.15}\\
         \bottomrule
    \end{tabular}
    \caption{Initialization Evaluation on EuRoC. Bold font indicates the best result, underline indicates the second best result. We compare these methods under 2 different configuration: 4KF and 5KF, both including scale, ATE, gravity, and success rate.}
    \label{tab:Init_Compare_sup}
\end{table*}

\subsection{Trajectory Evaluation}
In Sec. 6.5, we  conducted a comparison of trajectory accuracy between OKVIS \cite{leutenegger-ijrr-2015-OKVIS}, VINS-Mono \cite{qin-tro-2018_VINS-Mono}, VINS-Fusion \cite{qin2019a_VINS_Fusion_Local}, OpenVINS \cite{geneva2020openvins}, HybVIO \cite{hybvio} and our XR-VIO algorithm on the Euroc \cite{Burri25012016-EuRoC} and ZJU-Sensetime \cite{jinyu2019survey} datasets. The overall accuracy of the XR-VIO algorithm surpassed these SOTA algorithms on both datasets. However, Euroc and ZJU-Sensetime datasets do not provide metrics for Apple ARKit \footnote{\url{https://developer.apple.com/documentation/arkit/}} and Google ARCore \footnote{\url{https://developers.google.com/ar/}},  two leading commercial AR software solutions with substantial influence in the XR field. To further compare with ARKit and ARCore, we utilized the ADVIO dataset ~\cite{cortes2018advio}.

\begin{table}[h]
  \centering
    \begin{tabular}{cccccc}
    \toprule
         Dataset & ARKit & ARCore & XR-VIO \\
    \midrule
    advio-01 & 2.466 & 8.294 & \textbf{2.039} \\
    advio-02 & 2.594 & \textbf{1.473} & 2.387 \\
    advio-03 & \textbf{1.293} & 2.694 & 1.488 \\
    advio-04 & 5.186 & 24.67 & \textbf{2.827} \\
    advio-05 & 1.631 & \textbf{1.027} & 1.213 \\
    advio-06 & 4.093 & 1.758 & \textbf{1.499} \\
    advio-07 & 3.511 & 7.867 & \textbf{0.581} \\
    advio-08 & \textbf{1.318} & 7.018 & 1.540 \\
    advio-09 & 2.971 & \textbf{2.071} & 3.072\\
    advio-10 & 1.850 & \textbf{0.990} & 2.097 \\
    advio-11 & 2.782 & 2.822 & \textbf{2.677} \\
    advio-12 & 1.850 & 2.975 &  \textbf{1.719}\\
    advio-13 & 1.270 & 3.717 & \textbf{0.850} \\
    advio-14 & \textbf{1.500} & 5.825 & 4.990 \\
    advio-15 & 0.917 & 1.215 & \textbf{0.903} \\
    advio-16 & 1.219 & 3.069 & \textbf{0.576} \\
    advio-17 & 1.771 & 1.586 & \textbf{0.765} \\
    advio-18 & 0.695 & 1.509 & \textbf{0.589} \\
    advio-19 & 0.994 & \textbf{0.890} & 2.008 \\
    advio-20 & \textbf{9.445} & 12.77 & 12.931 \\
    advio-21 & 17.00 & 12.86 & \textbf{11.363} \\
    advio-22 & \textbf{4.752} & 5.198 & 6.467 \\
    advio-23 & -     & 4.660 & \textbf{4.252} \\
    \midrule
    \textbf{avg.} & 3.706 & 5.086 & \textbf{2.993} \\
    \bottomrule
    \end{tabular}%
  \caption{ATE (m) of different algorithms on the ADVIO dataset with metric of RMSE. Bold font indicates the best result in each column. ’-’ represents a failure to run on
this data.}
  \label{tab:data_sup}%
\end{table}%
\textbf{ADVIO} (Advanced Visual-Inertial Odometry) dataset is an open benchmark dataset designed for evaluating visual-inertia algorithms. It contains diverse real-world scenes, including different indoor/outdoor environments, lighting conditions, and dynamic object interferences, facilitating the assessment of robustness, accuracy, and real-time performance of various algorithms. Notably, the ADVIO dataset offers three aligned trajectories with ground truth: an ARCore trajectory captured on a Google Pixel device, an ARKit trajectory obtained from an iPhone, and Tango odometry recorded on a Google Tango tablet device.

Following the configurations outlined in Sec. 6, we executed all 23 datasets from ADVIO and compared them with ARKit and ARCore using Absolute Trajectory Error (ATE) metrics. As depicted in \cref{tab:data_sup}, out of the 23 datasets, our XR-VIO algorithm achieved the highest accuracy in 13 of them. Overall, our accuracy surpassed that of ARKit and ARCore. This experiment underscores the robustness and superiority of our XR-VIO algorithm in XR environments, particularly in pedestrian XR scenarios.

% The authors wish to thank A, B, and C. This work was supported in part by
% a grant from XYZ.}

%\bibliographystyle{abbrv}
\bibliographystyle{abbrv_journal/abbrv-doi}
\newpage
\bibliography{reference/main}